\documentclass[10pt,twocolumn,letterpaper]{article}

\usepackage{cvpr}              

\usepackage[dvipsnames]{xcolor}

\usepackage[utf8]{inputenc} 
\usepackage[T1]{fontenc}    
\usepackage{url}            
\usepackage{booktabs}       
\usepackage{amsfonts}       
\usepackage{nicefrac}       
\usepackage{microtype}      
\usepackage{xcolor}  

\usepackage{times}
\usepackage{latexsym}
\usepackage{tikz}
\usetikzlibrary{shapes,decorations}
\usepackage{amsmath,amssymb, bbm}
\usepackage{pifont}
\usepackage{mathtools}
\usepackage{caption}
\usepackage{subcaption}
\usepackage{tabularx}

\usepackage{graphicx}
\usepackage{multicol}
\usepackage{multirow}
\usepackage{booktabs} 
\def\savelastnode{\pgfextra\edef\tmpA{\tikzlastnode}\endpgfextra}
\def\restorelastnode{\pgfextra\edef\tikzlastnode{\tmpA}\endpgfextra}
\tikzstyle{mybox} = [draw=black, fill=yellow!20, thick,
    rectangle, rounded corners, inner sep=10pt, inner ysep=20pt]
\tikzstyle{mynewbox} = [draw=black, fill=yellow!20, thick,
    rectangle, rounded corners, inner sep=5pt, inner ysep=3pt]
\tikzstyle{fancytitle} =[fill=black, text=white]
\tikzstyle{title} = [append after command={%
    \savelastnode node[fancytitle,right=10pt] at (\tikzlastnode.north west)%
    {\large{#1}}\restorelastnode}]

\newcommand{\bluebold}[1]{\textcolor{Blue}{\large{\textbf{#1}}}}
\newcommand{\smallbluebold}[1]{\textcolor{Blue}{\scriptsize{\textbf{#1}}}}
\usepackage{soul}

\newcommand{\illusion}[1]{%
  \begingroup
  \sethlcolor{yellow}%
  \textcolor{red}{\hl{#1}}%
  \endgroup
}

\newcommand{\illusionunsure}[1]{%
  \begingroup
  \sethlcolor{yellow}%
  \textcolor{blue}{\hl{#1}}%
  \endgroup
}

\newcommand{\hallu}[1]{%
  \begingroup
  \sethlcolor{pink}%
  \textcolor{red}{\hl{#1}}%
  \endgroup
}

\newcommand{\mixed}[1]{%
  \begingroup
  \sethlcolor{Dandelion}%
  \textcolor{red}{\hl{#1}}%
  \endgroup
}

\newcommand{\tian}[1]{\textcolor{orange}{#1}}

\usepackage{soul}

\newcommand{\ours}{\textsc{HallusionBench}}

\usepackage{booktabs} 
\usepackage{makecell}

\usepackage{afterpage}
\usepackage{floatpag}

\definecolor{cvprblue}{rgb}{0.21,0.49,0.74}
\usepackage[pagebackref,breaklinks,colorlinks,citecolor=cvprblue]{hyperref}

\def\paperID{15855} 
\def\confName{CVPR}
\def\confYear{2024}

\title{\ours{}: An Advanced Diagnostic Suite for Entangled Language Hallucination and Visual Illusion in Large Vision-Language Models}

\author{
Tianrui Guan\thanks{Equal contribution.} 
~~~Fuxiao Liu\footnotemark[1] 
~~~Xiyang Wu
~~~Ruiqi Xian 
~~~Zongxia Li 
~~~Xiaoyu Liu 
~~~Xijun Wang \\
~~~Lichang Chen 
~~~Furong Huang
~~~Yaser Yacoob 
~~~Dinesh Manocha 
~~~Tianyi Zhou \\ \\
University of Maryland, College Park \\
\tt\small \texttt{\{rayguan, fl3es, wuxiyang, rxian, zli12321, xliu1231, xijun} \\
\tt\small \texttt{bobchen, furongh, yaser, dmanocha, tianyi\}@umd.edu} \\
}

\begin{document}
\maketitle

\begin{abstract}


We introduce ``\ours{}\footnote[1]{ ``Hallusion'' is a portmanteau of ``\textbf{hall}ucination'' and ``i\textbf{llusion}.''},'' a comprehensive benchmark designed for the evaluation of image-context reasoning. This benchmark presents significant challenges to advanced large visual-language models (LVLMs), such as GPT-4V(ision), Gemini Pro Vision, Claude 3, and LLaVA-1.5, by emphasizing nuanced understanding and interpretation of visual data.
The benchmark comprises 346 images paired with 1129 questions, all meticulously crafted by human experts.
We introduce a novel structure for these visual questions designed to establish control groups. This structure enables us to conduct a quantitative analysis of the models' response tendencies, logical consistency, and various failure modes.
In our evaluation on \ours{}, we benchmarked 15 different models, highlighting a 31.42\% question-pair accuracy achieved by the state-of-the-art GPT-4V. Notably, all other evaluated models achieve accuracy below 16\%.
Moreover, our analysis not only highlights the observed failure modes, including language hallucination and visual illusion but also deepens an understanding of these pitfalls. Our comprehensive case studies within \ours{} shed light on the challenges of hallucination and illusion in LVLMs. Based on these insights, we suggest potential pathways for their future improvement.
The benchmark and codebase can be accessed at \href{https://github.com/tianyi-lab/HallusionBench}{https://github.com/tianyi-lab/HallusionBench}.

\end{abstract}
\section{Introduction}


\begin{figure*}[h]
    \centering
      \includegraphics[width=\textwidth]{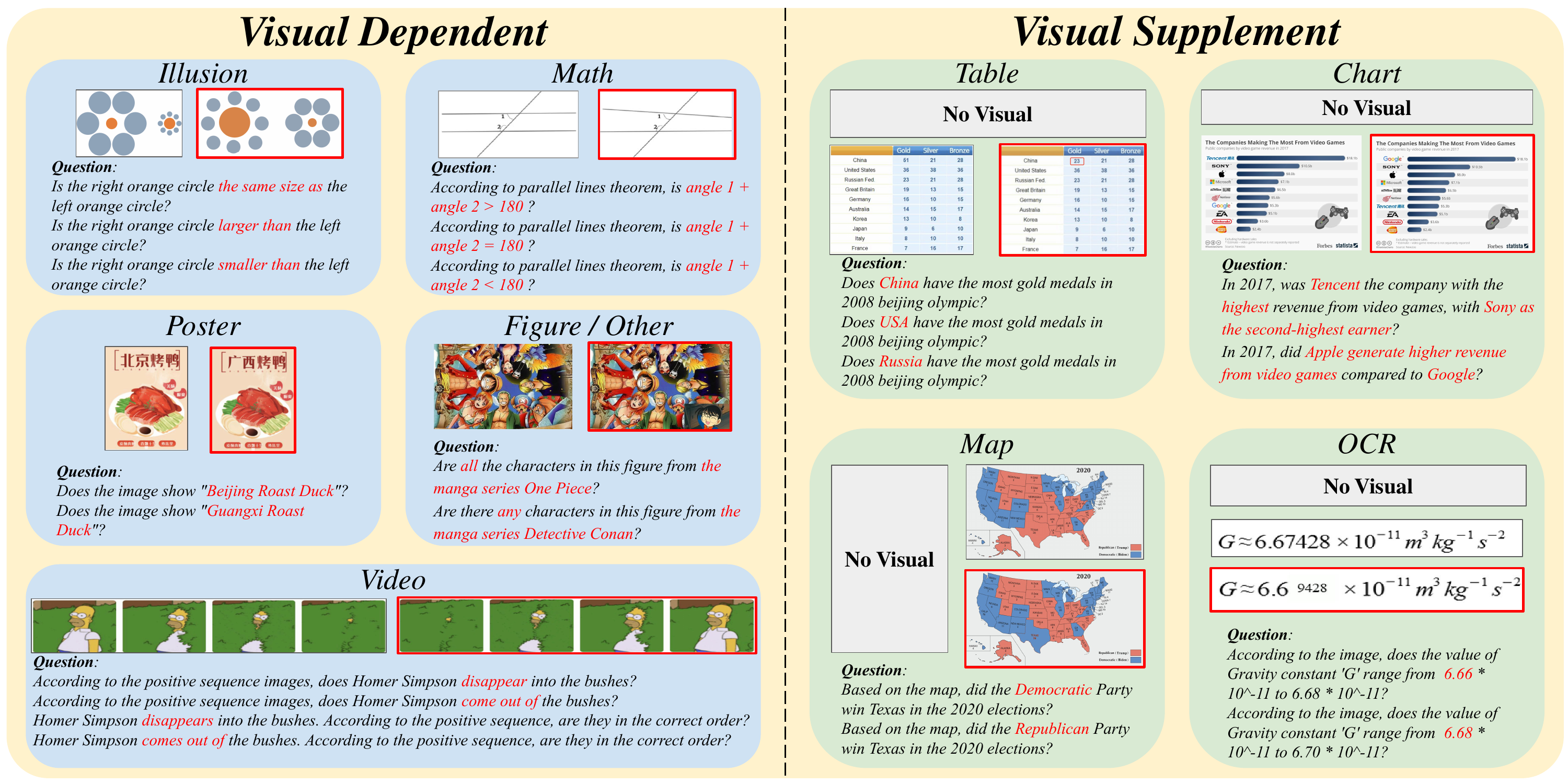}
    \caption{Data samples of \ours{}, which contains diverse topics, visual modalities. Human-edited images are in \textcolor{red}{RED}, resulting in different correct answers to the questions. 
    }
    \label{fig:examples}
\vspace{-5mm}
\end{figure*}

In recent years, Large Language Models (LLMs)~\citep{touvron2023llama,chiang2023vicuna,chen2023alpagasus,zhao2023survey,li2023towards,wei2023larger,Xiao2023LargeLM} have revolutionized the field of machine learning with the ability of language understanding and content generation, 
offering unprecedented capabilities and potentials across a multitude of applications. 
The integration of LLMs with computer vision systems has given rise to  Large Vision-Language Models (LVLMs)~\citep{bubeck2023sparks,touvron2023llama,zeng2022socratic,anas_awadalla_2023_7733589,yang2023mm,liu2023visual,zhu2023minigpt,ye2023mplug,liu2020visual, Li2023BLIP2BL, tran2020transform, liang2023modulewise}. 
 These models have demonstrated profound capabilities in various applications and significantly enhance the performance in image reasoning tasks~\citep{saikh2022scienceqa,hossain2019comprehensive,liu2023documentclip,vinyals2016show,liu2023covid,antol2015vqa,masry2022chartqa, guan2023loczson, yang2023embodied}. However, the hallucination issue of LLMs~\cite{Zhang2023SirensSI} is regarded as a challenging and unsolved problem, which leads to many issues when we integrate LLMs with vision techniques.

While LVLMs like GPT-4V(ision)~\cite{yang2023dawn} and LLaVA-1.5~\cite{liu2023improved} excel in various applications, they are hindered by a pronounced language bias. This bias stems from instances where knowledge priors conflict with the visual context~\cite{Li2023EvaluatingOH, liu2023aligning,zhai2023investigating}. Similarly, models such as LLaVA-1.5~\cite{liu2023improved} and mPLUG-Owl~\cite{ye2023mplug} are prone to giving affirmative answers regardless of the actual content of questions~\cite{Li2023EvaluatingOH}. The distinct failure modes of different VLMs highlight the need for specific improvements. Recognizing and understanding these limitations and failure types is imperative for advancing these models and striking a delicate balance between knowledge priors and contextual understanding.

When exploring those LVLMs, we observe that their strong language bias often overshadows visual information, leading to an overreliance on language priors rather than the visual context. To study this phenomenon, we use the term ``\textit{\textbf{Language Hallucination}},'' which refers to conclusions drawn without visual input.
On the other hand, the vision components within the limited ability in LVLMs can give rise to ``\textit{\textbf{Visual Illusion}}'', where visual inputs can be misinterpreted, leading to overconfident yet erroneous assertions by the model. 
\noindent\textbf{Main Contributions:}
Recognizing the need to comprehend why an LVLM fails and address these issues, we present \ours{}, a carefully crafted benchmark designed to explore the complexities of image-context reasoning in depth and expose various problems with respect to current LVLMs, as shown in Fig. \ref{fig:examples}. 
Our design of the visual-question (VQ) pairs, unique in format, facilitates a quantitative analysis of the models' failures, enabling a more thorough evaluation.
This investigation sheds light on existing limitations and lays the groundwork for future improvements, aiming to make the next generation of LVLMs more robust, balanced, and precise.
The novelties of our work include:
\begin{enumerate}
    \item We introduce \ours{}, the first advanced diagnostic suite
    tailored to systematically dissect and analyze
    the diverse failure modes of LVLMs.
    \ours{} consists of approximately 1129 handcrafted visual question-answer (VQA) pairs, 
    featuring 165 original images and 181 images expertly modified by human professionals. 
    Moving beyond the traditional metrics of correctness and accuracy, our VQA pairs are thoughtfully formulated with an innovative structure. This approach enables us to quantitatively analyze specific dimensions and aspects where current models falter.
    \item We evaluate 15 most recent methods on \ours{}. Our benchmark
    presents formidable challenges to existing methods. Notably, the SoTA GPT-4V achieves merely a 31.42\% Question Pair Accuracy, while the performance of all other methods falls below 16\%. 
    \item We explore \ours{} and provide an in-depth analysis of examples on which the SoTA LVLMs, such as GPT-4V and LLaVA-1.5 fail. We also provide insights on different issues that existing LVLMs are facing based on the quantitative analysis enabled by \ours{}.
    In our exploration of \ours{}, we conduct a detailed analysis of instances where SoTA LVLMs, including GPT-4V and LLaVA-1.5, fall short. Additionally, our investigation leverages the quantitative capabilities of \ours{} to shed light on various issues currently challenging existing LVLMs.


\end{enumerate}
\section{Related Work}

\subsection{Large Multi-Modal Models}
Large Language Models have been a major advancement, leading to new ways to understand not just text but other things like images, all in one large system. 
For example, Flamingo \cite{alayrac2022flamingo} has many capabilities, combining a vision part that doesn't change with a big language model that has a special feature for understanding both images and words together.
Another model, PaLM-E \cite{Driess2023PaLMEAE}, mixes visual information directly into the already powerful PaLM model, which has  $520$ billion parameters, making it effective in real-world uses. Most recently, researchers have been creating high-quality, diverse multi-modal datasets from GPT4 and GPT-4V \cite{yang2023dawn} to fine-tune open-source LVLMs, including LLaVA \cite{liu2023visual}, MiniGPT4 \cite{zhu2023minigpt}, Mplug-Owl \cite{ye2023mplug}, LRV-Instruction \cite{liu2023aligning}, LLaVAR \cite{Zhang2023LLaVAREV} and other works \cite{dai2023instructblip, peng2023instruction, Li2023StableLLaVAEV, yin2023survey}.

\subsection{Hallucination in LVLMs}
Hallucination typically refers to situations where the generated responses contain information that is not present in the visual content. Prior research primarily examines two areas: detecting and evaluating hallucinations \cite{Zhang2023SirensSI, Li2023EvaluatingOH, zhang2023grounding}, and methods to reduce them \cite{liu2023aligning, Yin2023WoodpeckerHC, Wang2023VIGCVI}. Early methods include training classifiers to identify hallucinations or comparing output with accurate answers to detect inaccuracies. To mitigate hallucinations, efforts have been made to improve data gathering and training procedures. For example, LRV-Instruction \cite{liu2023aligning} creates balanced positive and negative instructions to finetune LVLMs. VIGC \cite{Wang2023VIGCVI} uses an iterative process to generate concise answers and combine them, aiming for detailed yet accurate responses. Similarly, Woodpecker \cite{Yin2023WoodpeckerHC} introduces a training-free method to pick out and correct hallucinations from the generated text.

\subsection{Benchmarks for Large VL Models}
Traditional Visual Language (VL) benchmarks are designed to assess distinct skills, including visual recognition \cite{Goyal2016MakingTV}, image description \cite{Agrawal2019nocapsNO, liu2020visual}, and so on. However, with the advent of advanced LVLMs, 
traditional evaluation metrics often fall short of providing a detailed ability assessment. This problem is further exacerbated by their inability to match the given answer accurately, leading to significant robustness issues. To address these challenges, research communities have introduced a series of benchmarks, including MME \cite{fu2023mme}, MMBench \cite{Liu2023MMBenchIY}, MM-Vet \cite{Yu2023MMVetEL}, SEED-Bench \cite{Li2023SEEDBenchBM}, GAVIE \cite{liu2023aligning}, and LAMM-Bench \cite{Yin2023LAMMLM}. These benchmarks systematically structure and evaluate complex multi-modal tasks. Different from POPE \cite{Li2023EvaluatingOH} and GAVIE \cite{liu2023aligning} evaluating the object hallucinations of LVLMs, \ours{} is the first human-annotated analytical benchmark focusing on diagnosing both the visual illusion and knowledge hallucination of LVLMs.

\begin{figure*}[ht!]
  \begin{tabularx}{\linewidth}[t]{X*{9}{c}}
    \begin{tabular}[c]{p{\linewidth}}
      \centering
\setlength{\tabcolsep}{5pt} 
\renewcommand{\arraystretch}{1.2} 
\resizebox{0.75\columnwidth}{!}{%
\begin{tabular}{|c | l|c|c|c|c|c|}
\hline
\multicolumn{2}{|c|}{ }& \textbf{\thead{No\\Visual}} & 
\textbf{\thead{Original\\Visual} }
& \textbf{\thead{Edited\\Visual} }& \multicolumn{2}{c|}{\textbf{Overall}} \\
\hline
\multirow{5}{*}{{
    \begin{tabular}{c}
        \textbf{Visual} \\ \textbf{Dependent}\\
    \end{tabular}
    }} & \textit{Illusion} & - & 72  & 72  & 144 & \multirow{5}{*}{591} \\
                             & \textit{Math}    & - & 54  & 54  & 108 & \\
                             & \textit{Video}   & - & 69  & 101 & 170 & \\
                             & \textit{Poster}  & - & 43  & 46  & 89  & \\
                             & \textit{Others}  & - & 39  & 41  & 80  & \\
\hline
\multirow{4}{*}{{
    \begin{tabular}{c}
        \textbf{Visual} \\ \textbf{Supplement}\\
    \end{tabular}
    }} & \textit{Chart}   & 76  & 68  & 62  & 206 & \multirow{4}{*}{538} \\
                             & \textit{Table}   & 43  & 43  & 69  & 155 & \\
                             & \textit{Map}     & 32  & 32  & 32  & 96  & \\
                             & \textit{OCR}     & 27  & 27  & 27  & 81  & \\
\hline
\multicolumn{2}{|c|}{\textbf{Overall}} & 178 & 447 & 504 & \multicolumn{2}{c|}{1129} \\
\hline
\end{tabular}
}
    \end{tabular} 
&
    \centering
    \begin{tabular}[c]{c}
      \includegraphics[width=0.21\textwidth]{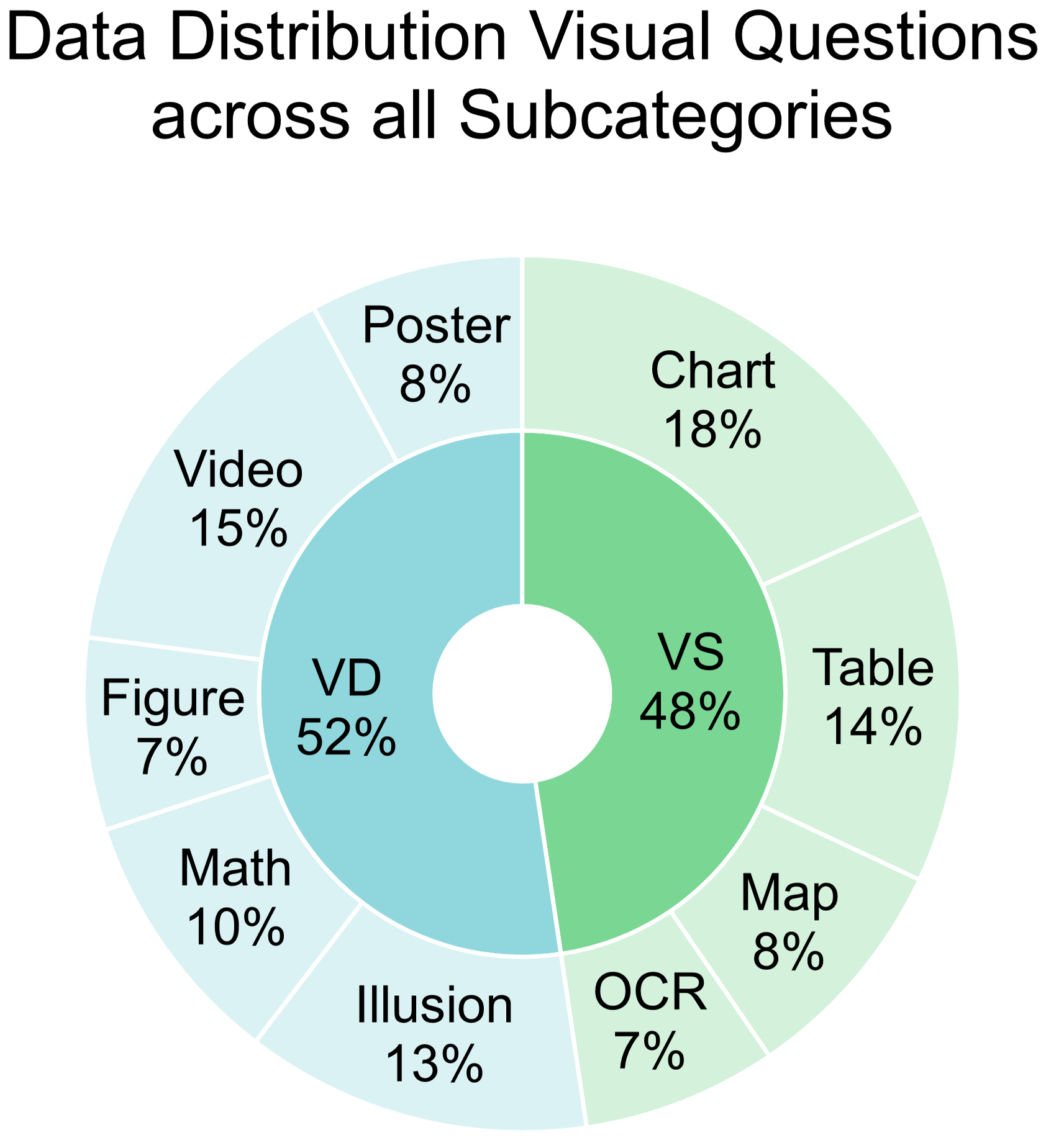}    \end{tabular} 
    &
        \centering
    \begin{tabular}[c]{c}
      \includegraphics[width=0.18\textwidth]{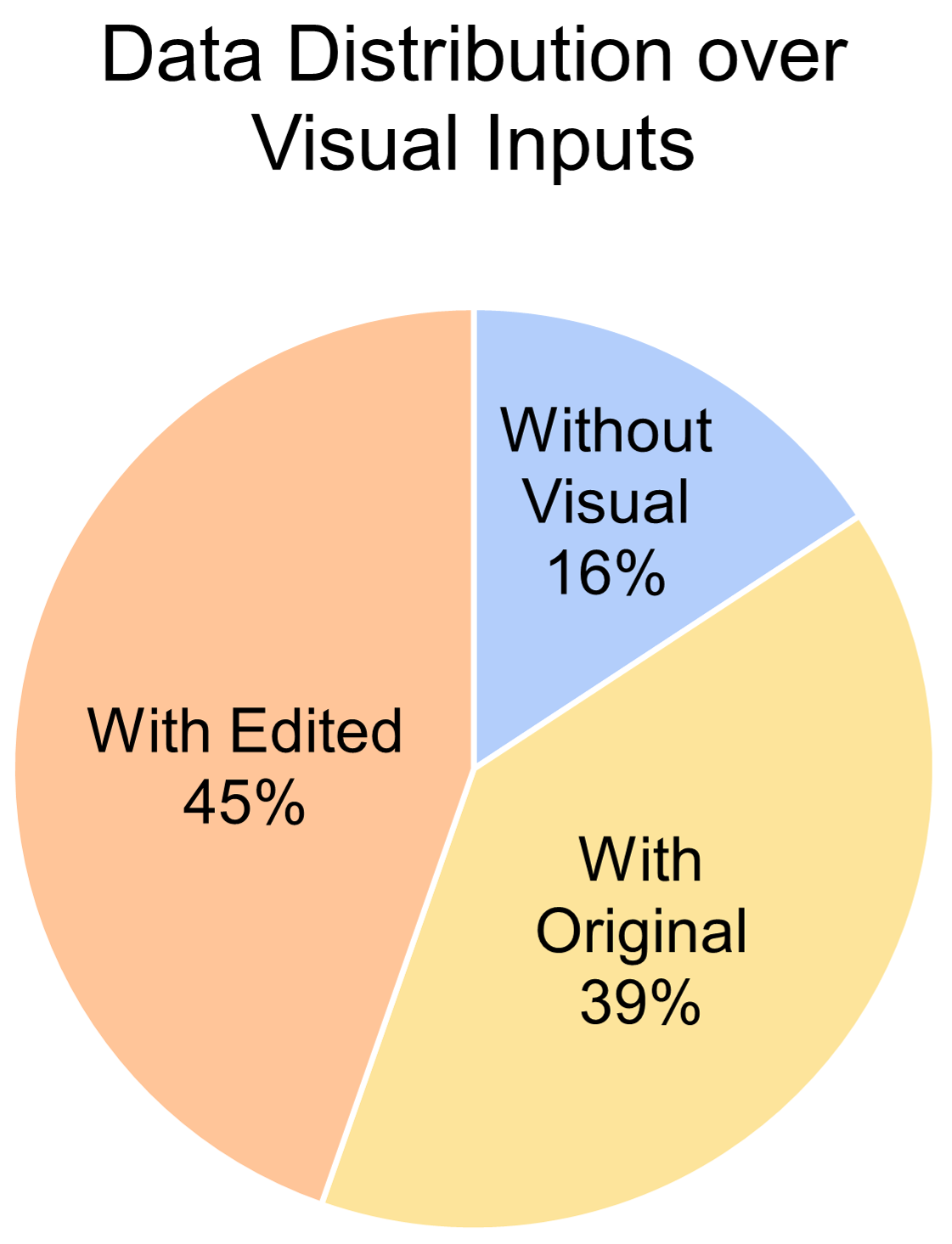}    \end{tabular}
\end{tabularx}
\vspace{-10pt}
\caption{\textbf{Statistics of \ours{}:} We show the number of questions in the table (\textit{left}), and the distribution of visual questions across each subcategory of Visual Dependent (VD) and Visual Supplement (VS) (\textit{middle}) and visual input types categorized by no visual, original, and edited images (\textit{right}). \ours{} covers a diverse visual format and nearly half of the images are manually edited.}
\vspace{-10pt}
\label{fig:stats}
\end{figure*}
\vspace{-2mm}
\section{\ours{} Construction}

\begin{table*}[h]
\setlength\tabcolsep{4.3pt}
\centering
\small
\resizebox{0.9\textwidth}{!}{
\begin{tabular}{l|ccccccc}
\toprule[1pt]
\textbf{Benchmarks} & \textbf{Visaul Format} & \textbf{\# Total QA} & \textbf{\# H-Edited QA} & \textbf{\# Total Img.} &\textbf{\# H-Edited Img.} & \textbf{Control Pair?} & \textbf{Purpose} \\
\midrule
Lynx-Bench~\cite{zeng2023matters}&  Image,Video 
& 450 & 450 & 450  &  0 & \ding{55}  & Image\&Video QA Evaluation \\
SciGraphQA~\cite{Li2023SciGraphQAAL} &  Image 
& 295K & 0 & 657K &  0 & \ding{55}  & Scientific Chart QA Evaluation \\
MathVista~\cite{Lu2023MathVistaEM} &  Image 
& 6141 & 0 & 5487 & 0  & \ding{55}  &   Math Reasoning Evaluation \\
MME~\cite{fu2023mme} &  Image 
& 1457  & 1457 & 1187 & 0 & \ding{55} & Comprehensive Evaluation \\
\midrule
POPE~\cite{Li2023EvaluatingOH} &  Image 
& 3000 & 0  & 500 &  0 &\ding{55} &  Object Hallucination \\
M-HalDetect~\cite{Gunjal2023DetectingAP} & Image 
& 4000 & 0  & 4000 &  0 &\ding{55} &   Object Hallucination \\
GAVIE~\cite{liu2023aligning} &  Image 
& 1000  &  0 & 1000  &  0 &\ding{55} &  Object Hallucination \\
Bingo~\cite{Cui2023HolisticAO} &  Image 
& 370 & 370  &  308 & N/A &\checkmark  &   Hallucination, Bias \\
\midrule
\ours{} &   {\begin{tabular}{c}
        Image, Video\\ \textbf{Image Pairs}
    \end{tabular}} &  1129 &1129&  346&    \textbf{181}& \checkmark  &
    {\begin{tabular}{c}
        \textbf{Visual Illusion,}\\ \textbf{Language Hallucination,}\\ \textbf{Quantitative Analysis and Diagnosis}\\
    \end{tabular}}\\

\bottomrule[1.5pt]
\end{tabular}
}
\vspace{-0.05in}
\caption{\textbf{Comparison of \ours{} with most recent VL benchmarks:}  
\ours{} is the \textbf{first} and the \textbf{only} benchmark that focuses on control-group analysis by carefully editing each image in the database manually. ``\# H-Edited QA'' means Human-edited question-answer pairs. ``\# H-Edited Img'' means Human-edited images. \textit{N/A} denotes that the information is not provided.
}
\label{tab:dataset_comparison}
\vspace{-0.20in}
\end{table*}

We present \ours{}, the first benchmark designed to examine visual illusion and knowledge hallucination of LVLMs and analyze the potential failure modes based on each hand-crafted example pair. \ours{} consists of 455 visual-question control pairs, including 346 different figures and a total of 1129 questions on diverse topics (including \ \textit{\textbf{food, math, geometry, statistics, geography, sports, cartoon, famous illusions, movie, meme,} etc.}) and formats (including \textit{\textbf{logo, poster, figure, charts, table, map, consecutive images,} etc.}). 
In the following sections, 
we first provide the guidelines for dataset construction based on different visual question types. Second, we will describe the data and annotation structure of \ours{}. Finally, we will describe the statistics of our dataset.
\vspace{-2mm}

\subsection{Visual Question Taxonomy}
\label{sec:question_type_def}

Our aim is to develop a multimodal image-context reasoning benchmark to investigate the potent language bias inherent in LVLMs, which can sometimes overshadow the visual context. We define the two categories of visual questions: \textit{\textbf{Visual Dependent}} and \textit{\textbf{Visual Supplement}}.
\vspace{-2mm}
\subsubsection{Visual Dependent Questions}
The \textbf{\textit{Visual Dependent}} questions are defined as \textbf{\textit{questions that do not have an affirmative answer without the visual context.}} Such questions ask about the image itself or something within the image. For example, there is no clear answer to \textit{"Is the right orange circle the same size as the left orange circle?"} without an image to provide more context.


\noindent\textbf{{Guideline}:} Under this setting, our benchmark is designed to evaluate visual commonsense knowledge and visual reasoning skills. Our exploration and dataset construction are guided by the following questions:

\begin{enumerate}
    \item \textit{How good are the visual understanding and reasoning skills of the model?}
    \item \textit{How does the parametric memory of the model affect its response to a question?}
    \item \textit{Is the model able to capture the temporal relation of multiple images?}
\end{enumerate}





\vspace{-2mm}
\subsubsection{Visual Supplement Questions}
The \textit{Visual Supplement} questions are \textbf{\textit{questions that can be answered without the visual input; the visual component merely provides supplemental information or corrections. }} For example, some LVLMs can answer \textit{"Is New Mexico state larger than Texas state?"} using the prior knowledge in their parametric memory without a map of the US. 

\noindent\textbf{\textit{Guideline}:} Under this setting, our benchmark is designed to evaluate visual reasoning ability and the balance between parametric memory and image context. Our exploration and dataset construction under this category is guided by the following questions:

\begin{enumerate}
    \item \textit{When the model lacks the prior knowledge or answer in the parametric memory of its language module, does the model (still) hallucinate about the images?}
    \item \textit{When the model's language module has sufficient prior knowledge in its parametric memory or directly knows the answer, does it still enhance its response by gathering extra information from the visual supplement (especially when the prior knowledge conflicts with the visual input or the parametric memory is outdated)?}
    \item \textit{How well can the model interpret a visual input with dense information (i.e., a graph, chart, map, etc.) for question answering? What types of image manipulation might impede or distort visual information extraction?} 
\end{enumerate}


\subsection{Visual, Question, and Annotation Structures}
\label{sec:structure}

\textbf{Notations:} Let $(I, q) \in \mathcal{V}\subseteq \mathbbm{I} \times \mathbbm{Q}$ be the tuple of the image $I\in \mathbbm{I}$ and question $q\in \mathbbm{Q}$, where $\mathcal{V}$ is the set of valid VQ pairs. 
Let $N$ be the number of original images obtained from the Internet, and $\mathbbm{I}_{o} = \{I_{(i, 0)}\}_{0 < i \leq N}$ be the set of those original images.
We define $\mathbbm{I}'_{i} = \{I_{(i,j)}\}_{0 < j \leq N_i}$ be the set of images modified from $I_{(i, 0)}$, and $I_0$ be an empty image. The entire images set $\mathbbm{I} = \{ I_0 \} \bigcup \mathbbm{I}_{o} \bigcup\ (\bigcup_{0 < i \leq N} \mathbbm{I}'_{i})$.


Let $\mathbbm{Q}_i = \{q_{(i,k)} \}_{{0 < k \leq M_i}}$ be the set of questions that can be applied to any image in $ \mathbbm{I}_i$, which is defined differently for Visual Dependent (\textit{VD}) and Visual Supplement (\textit{VS}):
\vspace{-3mm}
\begin{equation}
\mathbbm{I}_i = 
\begin{cases}
\{ I_{(i, 0)} \} \bigcup \mathbbm{I}'_i
& \text{for \textit{VD}}  \\
\{ I_0, I_{(i, 0)} \} \bigcup \mathbbm{I}'_i
& \text{for \textit{VS}}
\end{cases}
\label{eq:vdvs_cases}
\end{equation}
\vspace{-3mm}

To facilitate evaluation, all questions are formulated as Yes/No questions (Fig. \ref{fig:examples}). We annotate each visual-question with a binary answer $y(I, q) \in \{\text{``yes'', ``no''}\}$. 

\subsection{Dataset Statistics}
Following the annotation structure and guidelines above, we ask human experts to collect 346 images with diverse topics and types manually. As shown Fig.~\ref{fig:stats}, \textit{Visual Dependent} has 591 questions, including \textit{videos, illusion, math, posters, logos, cartoons}, and \textit{others}; \textit{Visual Supplement} has 538 questions, including \textit{charts, tables, maps}, and \textit{OCR}. Furthermore, Fig.~\ref{fig:stats} (\textit{right}) describes the distribution of the questions without visual input (16\%), with original online images (39\%), and with visual input edited by human experts (45\%). Our image manipulation strategies contain \textit{image flipping, order reversing, masking, optical character editing, object editing}, and \textit{color editing}. Additionally, each image has 3.26 questions on average. Fig.~\ref{fig:stats} (\textit{left}) provides more details on the number of questions in each topic and visual input category.




\subsection{Uniqueness of \ours{}}
The main comparison between \ours{} and existing benchmarks is presented in Tab. \ref{tab:dataset_comparison}. As it shows, there is a notable gap between existing benchmarks\cite{Li2023EvaluatingOH, Gunjal2023DetectingAP, liu2023aligning, Cui2023HolisticAO} and \ours{} in hallucination evaluation, as existing benchmarks primarily focus on object hallucinations, limited topics, and visual input types. Our dataset, \ours{}, is therefore motivated to bridge this gap by providing more topics, more image types, and more visual input modalities, including both images and videos. Additionally, our human experts carefully select each image and write question-answer pairs. We are also the first work to include human-edited images to assess the robustness of current LVLMs. Additionally, unlike existing benchmarks, \ours{} focuses on evaluating both language hallucinations and visual illusions, moving beyond the narrow scope of object hallucinations~\cite{Li2023EvaluatingOH, Gunjal2023DetectingAP, liu2023aligning}.

 \section{\ours{} Evaluation Suite}

\subsection{Text-Only GPT4-Assisted Evaluation}
\textbf{Notations:}  Let $\mathcal{M}(I, q)\in\{\text{``yes'', ``no'', ``uncertain''}\}$ be the parsed output answer by a VLM $\mathcal{M}$ for an image-question pair $(I, q)$. GPT-4 $GPT(\mathcal{M}(I, q),\ y(I, q))$ then judges the answer $\mathcal{M}(I, q)$ based on the ground truth $y(I, q)\in\{\text{``yes'', ``no''}\}$ and outputs \textit{Incorrect (0)}, \textit{Correct (1)}, or \textit{Uncertain (2)} if the predicted response is ambiguous.
The prompt for the GPT-4 judge is designed as:\looseness-1

\textit{Imagine you are an intelligent teacher. Thoroughly read the question, reference answer, and the prediction answer to ensure a clear understanding of the information provided. Assess the correctness of the predictions. If the prediction answer does not conflict with the reference answer, please generate “correct”. If the prediction answer conflicts with the reference answer, please generate “incorrect”. If the prediction answer is unclear about the answer, please generate "unclear".}
 
For each sample, we fill the template with its question, ground truth, and LVLM output. By taking the filled prompt into GPT-4, GPT-4 will generate "correct", "incorrect" or "unclear" for the sample. It is found that outputs of GPT-4 still exist variance, although the temperature is set as 0. Therefore, we utilize GPT-4 to evaluate the outputs of LLMs 3 times and report average scores.

\noindent\textbf{Comparison with Human Evaluation:} To demonstrate that our GPT4-Assisted evaluation is effective, we obtain the responses from GPT-4V~\citep{yang2023dawn} and LLaVA-1.5~\citep{liu2023improved}, and manually evaluate the correctness of their responses. We label the responses with  \textit{Incorrect (0)}, \textit{Correct (1)}, and \textit{Uncertain (2)} if the answer is ambiguous. As shown in the first two rows of Tab. \ref{tab:leaderboard} and Tab. \ref{tab:leaderboard-test}, the negligible difference proves that the GPT4-assisted method aligns well with human judgment.

\subsection{Correctness Evaluation Metrics}
Since the focus of our benchmark is on hallucination and illusion, not the span of knowledge, we consider an \textit{uncertain} answer acceptable when there is no visual input under the \textit{Visual Supplement} category. For the final accuracy score, we convert the correctness into a binary value $b_{\mathcal{M}}\in\{0, 1\}$:
\vspace{-1mm}
\begin{equation}
\resizebox{\columnwidth}{!}{$
b_{\mathcal{M}} (I, q) = 
\begin{cases}
GPT(\mathcal{M}(I, q),\ y(I, q))
& \text{if}\ GPT(\mathcal{M},\ y) \leq 1\\
1& \text{else if}\ I = I_{0} \\
0& \text{otherwise}
\end{cases},$}
\label{eq:correctness}
\end{equation}

Let $(I, q) \in \mathcal{V}\subseteq \mathbbm{I} \times \mathbbm{Q}$ be the tuple of the image $I\in \mathbbm{I}$ and question $q\in \mathbbm{Q}$, where $\mathcal{V}$ is the set of valid visual-question pairs.
Let $\mathbbm{1}(\cdot)$ be the indicator function. 

\noindent\textbf{All accuracy:} 
\vspace{-3mm}
\begin{equation}
     aAcc = \frac{\sum_{(I, q)\in \mathcal{V}} b_{\mathcal{M}} (I, q) }{|\mathcal{V}|}
     \label{metric:1}
\end{equation}
\vspace{-5mm}

\noindent\textbf{Figure Accuracy:}
\vspace{-1mm}
\begin{equation}
     fAcc = \frac{\sum_{i, j}{ \mathbbm{1}( \bigwedge_{q\in \mathbbm{Q}_i} b_{\mathcal{M}} (I_{(i,j)}, q)) }}{|\mathbbm{I}|}
     \label{metric:2}
\end{equation}
\vspace{-3mm}

\noindent\textbf{Question Pair Accuracy:}
\vspace{-1mm}
\begin{equation}
     qAcc = \frac{\sum_{i, k}{ \mathbbm{1}( \bigwedge_{I\in \mathbbm{I}_i} b_{\mathcal{M}} (I, q_{(i, k)})) }}{|\mathbbm{Q}|}     
     \label{metric:3}
\end{equation}
\vspace{-3mm}

\subsection{Analytical Evaluation Criteria}

In addition to the accuracy metrics, we introduce three analytical criteria to measure and diagnose the failures of LVLMs, \textit{Yes/No Bias Test}, \textit{Consistency Test}, and \textit{Diagnostic Test}. Instead of examining and analyzing each failed case qualitatively, we propose these novel quantitative measurements through the unique design of our question sets. These tests are listed in the order of complexity, so the latter test would not be as useful and insightful if the former basic test failed.

\subsubsection{Yes / No Bias Test} 

According to~\cite{Li2023EvaluatingOH}, some models~\cite{ye2023mplug, liu2023improved, gong2023multimodalgpt} tend to respond with ``yes'' in most cases. No further analysis is necessary if the model has a very strong bias or tendency to answer one way regardless of the actual question, so we design two criteria to reveal such preference of the model.

\noindent\textbf{Yes Percentage Difference (Pct. Diff) $d_{y}\in[-1, 1]$:} 
\vspace{-1mm}
\begin{equation}
     d_{y} = \frac{\sum\limits_{(I, q)\in\mathcal{V}}{ \left[\mathbbm{1}\big(\mathcal{M}(I, q)= \text{``yes''} \big) - \mathbbm{1}\big(y(I, q)=\text{``yes''} \big)\right]} }{ |\mathcal{V} |},
     \label{test:1}
\end{equation}
$d_{y}$ represents the difference between the predicted and actual number of ``Yes'' in the question set. The model is more biased when $|d_{y}|$ is close to 1.

\noindent\textbf{False Positive Ratio (FP Ratio) $r_{fp}\in[0, 1]$:}
\vspace{-3mm}
\begin{equation}
     r_{fp} = \frac{\sum_{(I, q)\in\mathcal{W}} \mathbbm{1}\big(\mathcal{M}(I, q)= \text{``yes"} \big)}{|\mathcal{W} |},    
     \label{test:2}
\end{equation}
where $\mathcal{W} = \{ (I, q) \in \mathcal{V}\ | \ b_{\mathcal{M}} (I, q) = 0\}$ is the set of incorrect visual questions. 
$r_{fp}$ measures how likely the model responses with ``Yes'' out of all incorrect responses. The model is more robust when $r_{fp}$ is close to 0.5.

\subsubsection{Consistency Test}

The goal of the consistency test is to test the logical consistency of responses and make sure questions are not answered based on random guesses.
Many questions $\mathbbm{Q}^i$ from root $\mathcal{R}^i$ are logically consistent: for example, ``Is the left segment longer than/shorter than/equal to the right segment?''
The consistency test is implemented and measured using \textit{fAcc} (Metrics~\ref{metric:2}). We design the question set $\mathbbm{Q}_i$ to be logically correlated over a figure. Therefore, we consider the model \textit{inconsistent} when only some of the questions in $\mathbbm{Q}_i$ are correct. In other cases, the model would be consistently correct or consistently wrong.


\subsubsection{Language Hallucination and Visual Illusion}
\label{sec:lhvi}

Before we dive into the diagnostic test, we categorize the failures into two major types based on the failed cases:

\textbf{Language Hallucination} \textit{refers to perceptions formed without relevant visual input.} In language hallucination, the model makes false prior assumptions about the input and image context based on its parametric memory. The model should respond based on how the question is framed instead of ignoring it or making false assumptions about the image.

\textbf{Visual Illusion} \textit{denotes the misinterpretation of accurate visual information.} Visual illusion comes from the failure to recognize and understand the input image visually. The model could not obtain accurate information or reason about the image correctly.

\begin{figure}[t]
    \centering
    \includegraphics[width=0.45\textwidth]{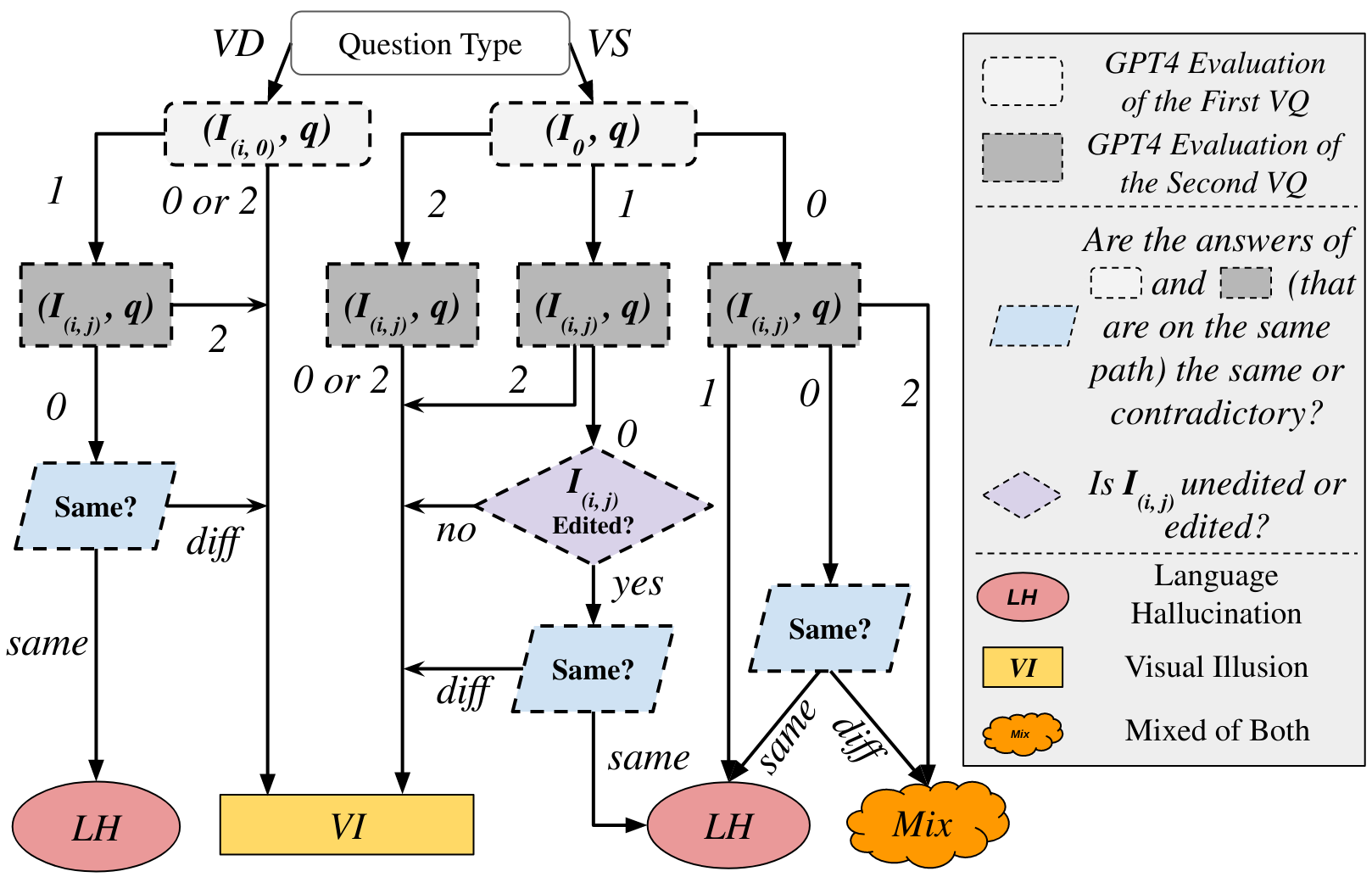}
    \caption{\textbf{Decision Tree to Diagnose Failure Types:} Based on the correctness of two questions in a control pair, and the difference of their responses, we use this decision tree to analyze the failure. The output of \textit{GPT4 Evalution} could be \textit{Incorrect (0)}, \textit{Correct (1)}, or \textit{Uncertain (2)} if the predicted response is ambiguous.}
    \label{fig:decisiontree}
    \vspace{-7mm}
\end{figure}

\subsubsection{Diagnostic Test}

To study the issue of language hallucination and language illusion, we analyze the responses and correctness of both visual questions within a \textit{VQ Control Pairs} 
and divide incorrect responses into three categories: \textit{Language Hallucination}, \textit{Visual Illusion}, and \textit{Mixed / Uncertain}. We measure the percentage of those failures out of all failed cases.

\noindent\textbf{Control Pair:} The control pair will always contain an original image for \textit{visual dependent} questions or an empty image (no visual) for \textit{visual supplement} questions. The other question in the control pair may have an edited image (or an original image for \textit{VS} question). The response to this question would provide more information on whether the answer exists in the parametric knowledge or if the model has seen it in the training data. In addition, we can examine whether the response remains the same after editing the original image to obtain more insights into the failures, which is more informative than checking a single visual question alone. In Fig.~\ref{fig:decisiontree}, we provide a decision tree to determine the type of failure for a control pair.
We consider the following principles when assigning the failure types:
\begin{enumerate}
    \item For \textit{visual dependent (VD)} questions, or \textit{visual supplement (VS)} questions that have visual inputs, if the response is incorrect or uncertain, the failure could be \textbf{\textit{visual illusion}}, since the model could not extract from the visual information correctly.
    \item For \textit{visual supplement (VS)} questions that don't have visual inputs, if the response gives a certain but wrong answer, we attribute it to \textbf{\textit{language hallucination}}.
    \item If the model responds to the original image (or no image) correctly and has the same response to the edited image (which is contrary to common sense), it means that the parametric knowledge overtakes the actual image input. Therefore, we also attribute the failure to \textbf{\textit{language hallucination}}.
\end{enumerate}
We will include some examples in the supplemental material.

\begin{table*}[t]
\vspace{-1em}
  \begin{center}
  \resizebox{1\textwidth}{!}{
\setlength{\tabcolsep}{4pt} 
  \begin{tabular}{lccccccc}
    \toprule            
    \textbf{Method} & \textbf{\# Parameter} & \textbf{Evaluation} & {
    \begin{tabular}{c}
        \textbf{Question Pair Accuracy} \\ \textbf{} (\textit{qAcc}) $\uparrow$\\
    \end{tabular}
    }
    & {
    \begin{tabular}{c}
        \textbf{Figure Accuracy} \\ (\textit{fAcc}) $\uparrow$\\
    \end{tabular}
    }
     & {
    \begin{tabular}{c}
        \textbf{Easy Accuracy} \\ (Easy \textit{aAcc}) $\uparrow$\\
    \end{tabular}
    }
    & {
    \begin{tabular}{c}
        \textbf{Hard Accuracy} \\ (Hard \textit{aAcc}) $\uparrow$\\
    \end{tabular}
    }
    & {
    \begin{tabular}{c}
        \textbf{All Accuracy} \\ (\textit{aAcc}) $\uparrow$\\
    \end{tabular}
    } \\ 
    \midrule
     \multirow{2}{*}{GPT4V~\cite{2023GPT4VisionSC} (Oct 2023)} & \multirow{2}{*}{-} & Human  &  31.42  &       44.22      &   79.56   &     38.37    &     67.58  \\
     & & GPT4-Assisted  &  \textbf{28.79}  &     \textbf{39.88}   &    \textbf{75.60}  &              37.67    &  \textbf{65.28}     \\
     \midrule
     \multirow{2}{*}{LLaVA-1.5~\cite{liu2023improved}} & \multirow{2}{*}{13B} & Human  &  9.45  &        25.43        &          50.77    &     29.07   &       47.12   \\
    &  & GPT4-Assisted & 10.55  &  \textbf{24.86} &  \textbf{49.67}   &           29.77    &  46.94  \\
    \midrule
        Claude 3~\cite{claude3} & -
     & GPT4-Assisted &    \textbf{21.76} & \textbf{28.61} & \textbf{55.16} & \textbf{41.40} & \textbf{56.86}\\   
     \midrule
    \hspace{-2mm}{
    \begin{tabular}{l}
        Gemini Pro Vision~\cite{google_gemini} \\ (Dec 2023)\\
    \end{tabular}
    }
    & -
     & GPT4-Assisted &    7.69 & 8.67 & 35.60 & 30.23 & 36.85    \\       
     \midrule
    BLIP2-T5~\cite{Li2023BLIP2BL} & 12.1B
     & GPT4-Assisted &    \textbf{15.16}     &       20.52     &     45.49           &   \textbf{43.49}    &         \textbf{48.09}     \\   
     \midrule
     Qwen-VL~\cite{Bai2023QwenVLAF} & 9.6B & GPT4-Assisted   &    5.93   &    6.65   &      31.43           &      24.88    &   39.15     \\
     Open-Flamingo~\cite{alayrac2022flamingo} & 9B
     & GPT4-Assisted &    6.37     &      11.27     &     39.56              &    27.21    &         38.44     \\
     MiniGPT5~\cite{Zheng2023MiniGPT5IV} & 8.2B & GPT4-Assisted &   10.55  &      9.83   &     36.04    &            28.37    &    40.30    \\
     MiniGPT4~\cite{zhu2023minigpt} & 8.2B & GPT4-Assisted &    8.79   &  10.12   &  31.87             & 27.67   & 35.78      \\
     InstructBLIP~\cite{dai2023instructblip} & 8.2B & GPT4-Assisted  &  9.45   & 10.11  &       35.60      & \textbf{45.12}   &     45.26    \\
     BLIP2~\cite{Li2023BLIP2BL} & 8.2B
     & GPT4-Assisted &    5.05     &       12.43     &      33.85              &    40.70    &         40.48     \\   
     mPLUG\_Owl-v2~\cite{ye2023mplugowl2} & 8.2B & GPT4-Assisted &    13.85    &      19.94      &   44.84    &    39.07    &  47.30      \\
     mPLUG\_Owl-v1~\cite{ye2023mplug} & 7.2B & GPT4-Assisted &   9.45    &     10.40   &       39.34             &       29.77   &     43.93      \\
     LRV\_Instruction~\cite{liu2023aligning} & 7.2B 
     & GPT4-Assisted  & 8.79   &    13.01     &            39.78             &           27.44     &     42.78        \\
      \midrule
     GIT~\cite{Wang2022GITAG} & 0.8B & GPT4-Assisted  &  5.27   &    6.36   &    26.81   &            31.86  &    34.37  \\
     \midrule
     Random Chance & - & GPT4-Assisted  &  15.60   &    18.21   &    39.12   &            39.06  &    45.96  \\
  \bottomrule
\end{tabular}
}
\vspace{-6mm}
\end{center}
\caption{\textbf{Correctness Leaderboard on \ours{} with various LVLMs:} All the numbers are presented in \% and the full score is 100\%. Hard questions refer to the edited images. We highlight the Top 3 models with the GPT4-assisted evaluation. }
\label{tab:leaderboard}
\end{table*}

\begin{table*}[t]
\vspace{-1em}
  \begin{center}
  \resizebox{1\textwidth}{!}{
\setlength{\tabcolsep}{4pt} 
  \begin{tabular}{lcccccccccc}
    \toprule
    \multicolumn{3}{c}{} &  \multicolumn{2}{c}{\textbf{Yes/No Bias}} &  \multicolumn{3}{c}{\textbf{Consistency}} &  \multicolumn{3}{c}{{ \textbf{Language and Vision Diagnosis}
    }}  \\ \cmidrule(lr){4-5} \cmidrule(lr){6-8}  \cmidrule(lr){9-11}              
    \textbf{Method} & \textbf{\# Parameter} & \textbf{Evaluation} & Pct. Diff ($\sim 0$) & FP Ratio ($\sim0.5$)& Correct $\uparrow$
    & Inconsistent $\downarrow$ & Wrong $\uparrow$ & Language Hallucination & Visual Illusion & Mixed\\
    \midrule
    \multirow{2}{*}{GPT4V~\cite{2023GPT4VisionSC} (Oct 2023)} &  \multirow{2}{*}{-} & Human  & 0.066 & 0.60 & 44.22 & 32.66 & 23.12 & 21.86 & 46.17 & 31.97   \\
    & & GPT4-Assisted  & \textbf{0.058} & 0.58 & \textbf{39.88} & \textbf{38.15} & \textbf{21.97} & 22.19 & 45.66 & 32.14   \\
     \midrule
    \multirow{2}{*}{LLaVA-1.5~\cite{liu2023improved}} &  \multirow{2}{*}{13B} & Human  & 0.27 &   0.76   &          25.43      &     42.49   &         32.08          & 25.63 & 51.42 & 22.95  \\
    & & GPT4-Assisted & 0.26   &   0.75   &          \textbf{24.86}        &             \textbf{45.38}          &         29.77          & 26.71 & 51.09 & 22.20 \\
    \midrule 
        Claude 3~\cite{claude3} & -
     & GPT4-Assisted &  0.063 & \textbf{0.57} & \textbf{28.61} & \textbf{49.42} & \textbf{21.97} & 19.10 & 59.14 & 21.77\\   
     \midrule
    \hspace{-3mm}{
    \begin{tabular}{l}
        Gemini Pro Vision~\cite{google_gemini} \\ (Dec 2023)\\
    \end{tabular}
    }
    & -   & GPT4-Assisted & \textbf{-0.02}  &  \textbf{0.48}   &  8.67 & 56.94  & 34.39 & 25.95 & 49.37 & 24.68 \\
     \midrule
    BLIP2-T5~\cite{Li2023BLIP2BL} & 12.1B & GPT4-Assisted & 0.08 & 0.58 & 20.52 & 59.54 & 19.94 & 41.64 & 40.44 & 17.92\\
     \midrule
    Qwen-VL~\cite{Bai2023QwenVLAF} & 9.6B & GPT4-Assisted & 0.12 & 0.60 & 6.65 & 50.29 & 43.06 & 0.87 & 88.06 & 11.06\\
    Open-Flamingo~\cite{alayrac2022flamingo} & 9B  & GPT4-Assisted & 0.33 & 0.77 & 11.27 & 59.83 & 28.90 & 30.07 & 48.06 & 21.87\\ 
    MiniGPT5~\cite{Zheng2023MiniGPT5IV} & 8.2B & GPT4-Assisted & 0.28   &   0.71   &           9.83           &             56.36             &          33.82          & 10.09 & 73.44 & 16.47 \\
    MiniGPT4~\cite{zhu2023minigpt} & 8.2B & GPT4-Assisted & 0.19 & 0.65 & 10.12 & 57.80 & 32.08 & 23.59 & 56.55 & 19.86\\
    InstructBLIP~\cite{dai2023instructblip} & 8.2B & GPT4-Assisted & -0.13 & 0.38 & 10.12 & 68.50 & \textbf{21.39} & 29.29 & 54.53 & 16.18\\
    BLIP2~\cite{Li2023BLIP2BL} & 8.2B  & GPT4-Assisted & 0.18 & 0.65 & 12.43 & 63.01 & 24.57 & 39.14 & 43.45 & 17.41\\
    mPLUG\_Owl-v2~\cite{ye2023mplugowl2} & 8.2B & GPT4-Assisted & 0.25 & 0.77 & 19.94 & 58.09 & 21.97 & 28.24 & 50.42 & 21.34\\
    mPLUG\_Owl-v1~\cite{ye2023mplug} & 7.2B & GPT4-Assisted & 0.32 & 0.79 & 10.40 & 60.12 & 29.48 & 3.95 & 78.36 & 17.69\\
    LRV\_Instruction~\cite{liu2023aligning} & 7.2B  & GPT4-Assisted & 0.26 & 0.73 & 13.01 & 53.47 & 33.53 & 4.49 & 76.47 & 19.04\\
     \midrule
    GIT~\cite{Wang2022GITAG} & 0.8B & GPT4-Assisted & \textbf{0.04} & \textbf{0.53} & 6.36 & 53.76 & 39.88 & 30.90 & 58.30 & 10.80 \\
    \midrule
    Random Chance & - & GPT4-Assisted & 0.08 & 0.57 & 18.20 & 57.51 & 24.28 & - & - & - \\

  \bottomrule
\end{tabular}
}
\end{center}
\vspace{-6mm}
\caption{\textbf{Analytical Evaluation Results on \ours{} with various LVLMs:} \textit{Pct. Diff} ranges from [-1, 1]. The model is more biased when \textit{Pct. Diff} is close to -1 or 1. \textit{FP Ratio} ranges from [0, 1]. The model is more robust when \textit{FP Ratio} is close to 0.5. All the other metrics are presented in \%, and the full score is 100\%. We highlight the Top 3 models with the GPT4-assisted evaluation. }
\label{tab:leaderboard-test}
\vspace{-5mm}
\end{table*}

\vspace{-2mm}

\section{Experimental Results}

\subsection{Models}

We conduct massive experiments on \ours{} to evaluate a total of 15 LVLMs, including GPT-4V \cite{2023GPT4VisionSC}, LLaVA-1.5 \cite{liu2023improved}, Gemini Pro Vision \cite{google_gemini}, Claude 3~\cite{claude3}, MiniGPT4 \cite{zhu2023minigpt}, MiniGPT5 \cite{Zheng2023MiniGPT5IV}, GiT \cite{Wang2022GITAG}, InstructBLIP \cite{dai2023instructblip}, Qwen-VL \cite{Bai2023QwenVLAF}, mPLUG-Owl-v1 \cite{ye2023mplug}, mPLUG-Owl-v2 \cite{ye2023mplugowl2}, LRV-Instruction \cite{liu2023aligning}, BLIP2 \cite{Li2023BLIP2BL}, BLIP2-T5 \cite{Li2023BLIP2BL}, and Open-Flamingo \cite{alayrac2022flamingo}. We also include \textit{Random Chance} (i.e. randomly choose \textit{Yes} or \textit{No}) as a baseline.

\subsection{Result Analysis}
\begin{figure}[t]
    \centering
    \vspace{-4mm}
      \includegraphics[width=0.37\textwidth]{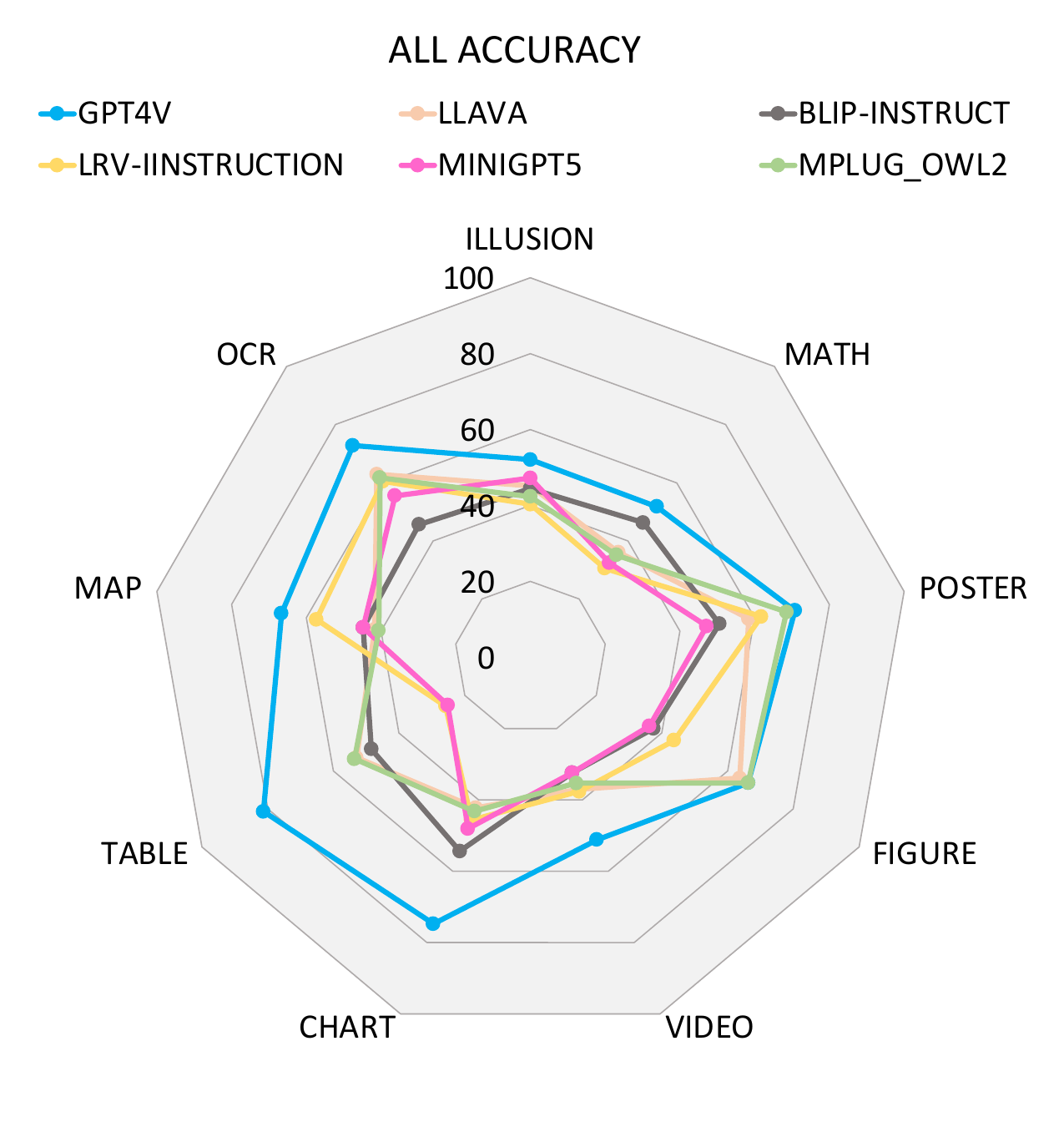}
    \vspace{-6mm}
    \caption{\textbf{Accuracies on each subcategories:} We show six prominent LVLMs on \ours{} across different types. }
    \label{fig:leidatu}
    \vspace{-7mm}
\end{figure}

We compare the performance of several models, including both closed-source models and open-sourced models. Results are given in Tab. \ref{tab:leaderboard}, Tab. \ref{tab:leaderboard-test} and Fig. \ref{fig:leidatu}. Additionally, we established a human expert evaluation to assess the effectiveness of text-only GPT4-assisted evaluation.  

\textbf{Correctness Evaluation. }As shown in Tab. \ref{tab:leaderboard}, GPT-4V outperforms all the open-sourced LVLMs by a large margin except the \textit{Hard Accuracy}. \textit{Hard Accuracy} measures the models' ability to understand human-edited images from \ours{}. The poor accuracy demonstrates the challenges of our image manipulations for GPT-4V and other open-source LVLMs. In the open-sourced models, we investigate if expanding the size (0.8B to 13B) of the LLM backbone can mitigate object existence hallucination. As detailed in Tab. \ref{tab:leaderboard}, there is a noticeable reduction in hallucination as the model size increases, like LLaVA-1.5 and BLIP2-T5. Among models with a size of less than 10B, InstructBLIP and mPLUG-Owl-v2 are the best-performing ones. InstructBLIP, leveraging the BLIP-2 architecture and enhanced through instruction fine-tuning across 26 diverse datasets, demonstrates that a broader and more extensive training set can substantially enhance performance. The boosting performance of mPLUG-Owl-v2 compared with mPLUG-Owl-v1 can be attributed to its novel module, which utilizes the language decoder acting as a universal interface for managing different modalities.

\textbf{Yes/No Bias. }Another observation is that GPT-4V, BLIP2-T5, and mPLUG-Owl-v2 outperform \textit{Random Choice} in both question pair accuracy, figure pair accuracy, and question level accuracy. Other models, such as Qwen-VL and MiniGPT4, perform even worse than \textit{Random Choice}. This indicates their visual reasoning abilities are still limited. However, LLaVA-1.5 outperforms \textit{Random Choice} while achieving poor results in both question pair accuracy and figure pair accuracy. We attribute this phenomenon to the fact that LLaVA-1.5 tends to answer \textit{Yes}. This assumption is supported by the low \textit{Yes Percentage Difference} and \textit{False Positive Ratio} of LLaVA-1.5 in \textit{Yes/No Bias Test} from Tab. \ref{tab:leaderboard-test}. Besides, we find that Open-Flamingo and mPLUG-Owl-v1 also tend to answer \textit{Yes} with the high \textit{Yes Percentage Difference} and \textit{False Positive Ratio}. Inspired by \cite{liu2023aligning}, one possible reason is that these LVLMs lack balanced positive and negative instructions in their training set. We also attribute the poor performance of these LVLMs to the scarcity of human-edited images in their training set since most LVLMs only utilize original images from existing datasets. 
\vspace{-1mm}

\begin{figure}
\scriptsize
\begin{center}
\vspace{-4mm}
\begin{tikzpicture}
\node [mynewbox] (box){%
    \begin{minipage}{\columnwidth}
        \raggedright
        \smallbluebold{Prompt: } Is angleA + angleB +angleC = 180? \\[2pt]\vspace{-3mm}

    \begin{multicols}{2}

        \centering
        \includegraphics[width=0.5\columnwidth]{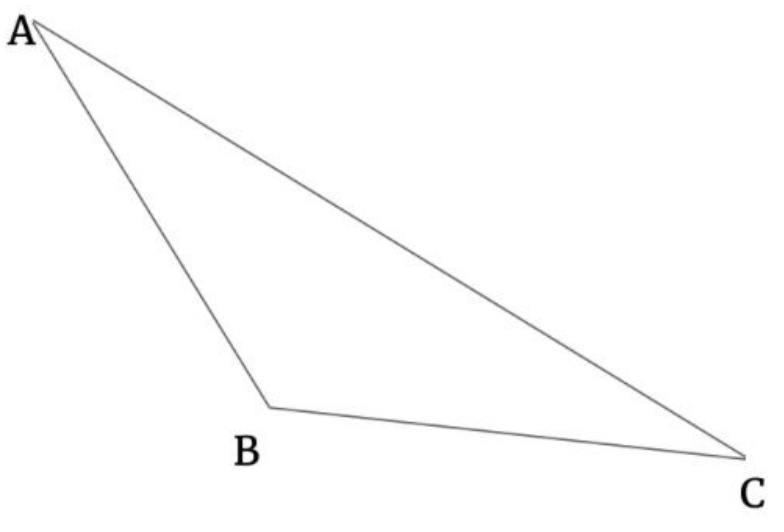}
        
        \textit{\textbf{Original image:} Yes.} \\ [2pt]
        \raggedright
        \smallbluebold{GPT-4V: } Yes, the sum of the interior angles of a triangle always equals 180 degrees. So, angleA + angleB +angleC = 180.\\[2pt]
        \smallbluebold{LLaVA-1.5: } Yes, angle A + angle B + angle C = 180.

        \columnbreak
    
        \centering
        \includegraphics[width=0.5\columnwidth]{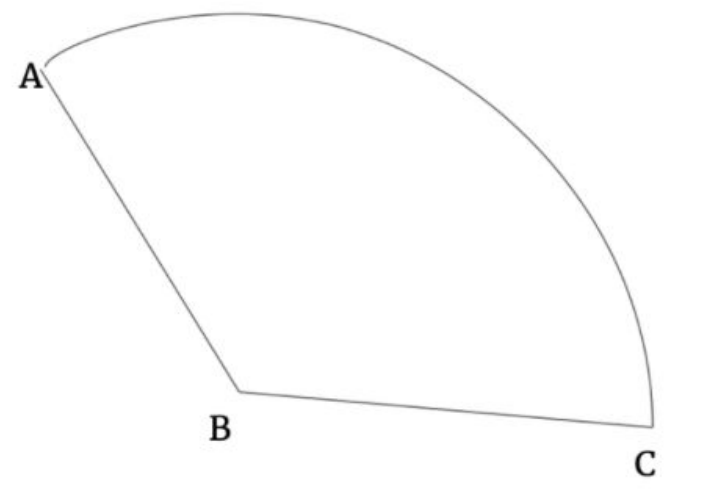}
        
        \textit{\textbf{Edited image:} No.} \\ [2pt]
        
        \raggedright
        \smallbluebold{GPT-4V: } Yes, the sum of the interior angles of a triangle always equals 180 degrees. So, \illusion{angleA + angleB +angleC = 180.} \\[2pt]
        \smallbluebold{LLaVA-1.5: } Yes, \illusion{angleA + angleB + angleC = 180.}

    \end{multicols}\vspace{-5mm}
        \par\noindent\rule[1pt]{\textwidth}{1pt}\\ 
        \raggedright
        \smallbluebold{Prompt: } Are the two yellow lines the same length?
 \\[2pt]\vspace{-3mm}

    \begin{multicols}{2}

        \centering
        \includegraphics[width=0.5\columnwidth]{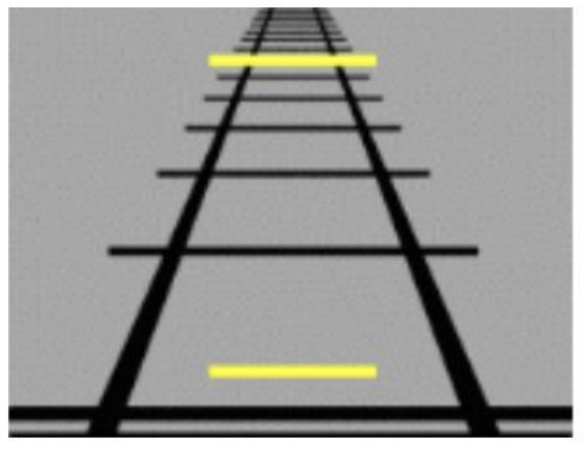}
        
        \textit{\textbf{Original:} Yes, the yellow lines have the same length.} \\ [2pt]
        \raggedright
        \smallbluebold{GPT-4V: } Yes, the two yellow lines are the same length. This is an example of an optical illusion where the surrounding context can make objects appear different in size or length than they actually are. \\[2pt]
        \smallbluebold{LLaVA-1.5: } No, \illusion{the two yellow lines are not the same length. }

        \columnbreak
    
        \centering
        \includegraphics[width=0.5\columnwidth]{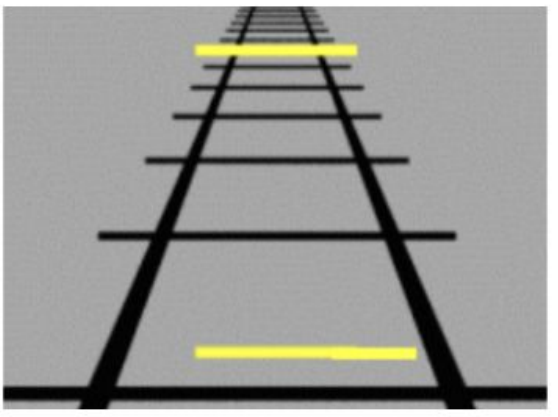}
        
        \textit{\textbf{Edited:} No, the bottom yellow line is elongated.} \\ [2pt]
        
        \raggedright
        \smallbluebold{GPT-4V: } Yes, the two yellow lines appear to be \hallu{the same length. This is an example of an optical illusion where parallel lines are in a converging context.} \\[2pt]
        \smallbluebold{LLaVA-1.5: } Yes, \illusion{the two yellow lines are the same length.}
        
    \end{multicols}\vspace{-5mm}
        \par\noindent\rule[1pt]{\textwidth}{1pt}\\ 
        \raggedright
        \smallbluebold{Prompt: } According to the positive sequence of the images,  does the baby attempt to put on the clothes? \\[2pt]\vspace{-3mm}

    \begin{multicols}{2}

        \centering
        \includegraphics[width=\columnwidth]{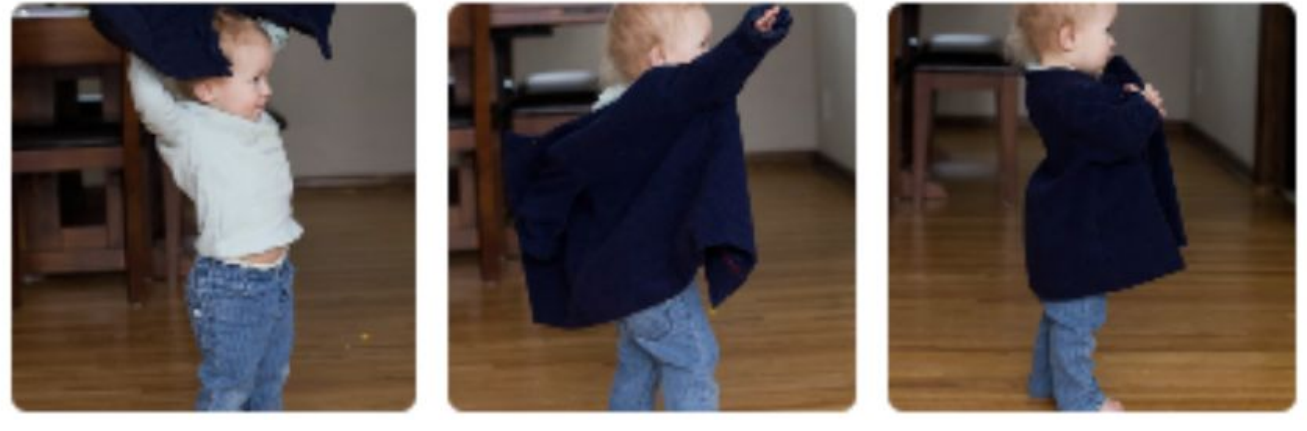}
        
        \textit{\textbf{Original sequence:} Yes.} \\
        \raggedright
        \smallbluebold{GPT-4V: } Yes, according to the sequence of the images, it appears that the baby is attempting to put on the clothes.
        
        \columnbreak
    
        \centering
        \includegraphics[width=\columnwidth]{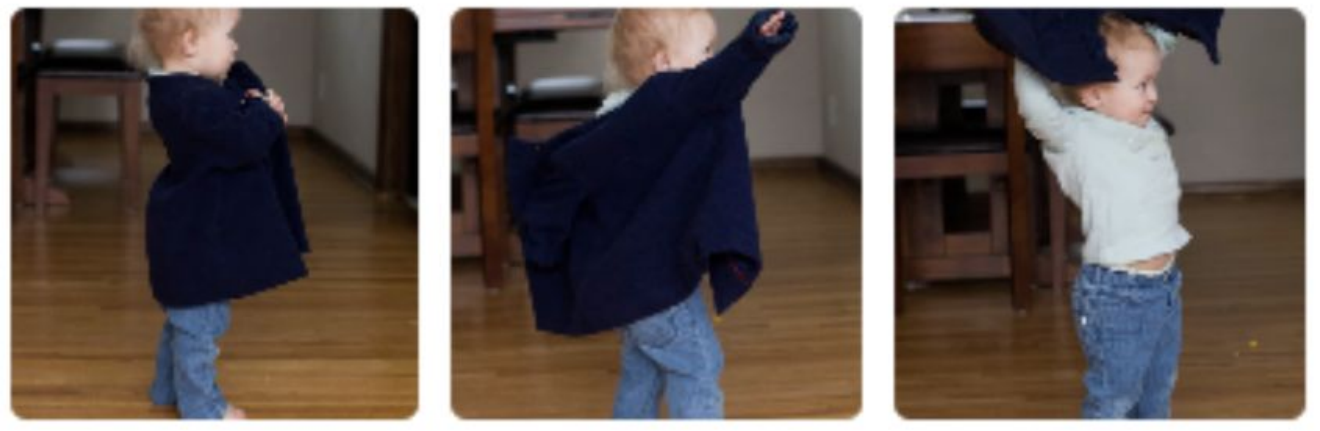}
        
        \textit{\textbf{Reversed sequence:} No.} \\
        
        \raggedright
        \smallbluebold{GPT-4V: } Yes, according to the sequence of the images, it appears that the baby is attempting to \illusion{put on the clothes}. \\
        
    \end{multicols}

    \end{minipage}
};
\end{tikzpicture}%
\vspace{-3mm}
\caption{
\textbf{Failure Cases in \textit{Math, Illusion and Video}:} We highlight \hallu{\textit{language hallucination}} and \illusion{\textit{visual illusion}}.
}
\label{fig:failurecase}
\vspace{-7mm}
\end{center}
\end{figure}

\textbf{Language and Vision Diagnosis. }We report fine-grained scores of six prominent LVLMs across different visual inputs in Fig. \ref{fig:leidatu}. Results show that \textit{Math}, \textit{Illusion}, and \textit{Video} is the most challenging format for current LVLMs, including GPT-4V. 
From Fig. \ref{fig:failurecase} (top), we found both GPT-4V and LLaVA-1.5 are unable to correctly recognize regular triangles, meaning that geometry and math are still a challenging task for GPT-4V. 
From Fig. \ref{fig:failurecase} (middle), we found GPT-4V is more knowledgeable than LLaVA-1.5 in recognizing all the illusion cases and knowing their names. However, GPT-4V fails to answer the question faithfully based on the edited images. The reason behind this might be that GPT-4V tends to generate answers based on its parametric memory instead of analyzing the images. Compared to GPT-4V, LLaVA-1.5 performs badly on both the original image and edited images, indicating that the visual perception skill of LLaVA-1.5 is limited. 
From Fig. \ref{fig:failurecase} (bottom), we found that GPT-4V is unable to distinguish between the positive sequence and the reversed sequence of the images, indicating that there is still much room to improve the video reasoning ability.

\section{Conclusion, Limitations and Future Work}
In this work, we introduce \ours{}, the first advanced diagnostic suite to analyze the failure cases of 15 current LVLMs.
\ours{} presents significant challenges to existing LVLMs like GPT-4V(ision), by emphasizing nuanced understanding and interpretation of visual data.
Moreover, our unique design of the visual-question pairs facilitates a quantitative analysis of the models' failures, enabling a more thorough evaluation.
We share our observations and key insights for future studies:
\begin{enumerate}
    \item When GPT-4V, LLaVA-1.5, and other LVLMs have prior knowledge of questions in \ours{}, they usually suffer from Language Hallucination as they tend to prioritize their prior knowledge which leads to incorrect answers. 
    The model should handle the trade-off between parametric memory and context.
    \item When LVLMs have not had parametric memory or prior knowledge regarding the questions in \ours{}, they can still be prone to Visual Illusion and prefer to produce wrong answers about the given figure.
    The visual capability of existing LVLMs is still limited.
    \item  GPT-4V and other LVLMs can be easily misled by simple image manipulations in \ours{}, including image flipping, order reversing, masking, optical character editing, object editing, and color editing. 
    \item GPT-4V and other LVLMs are unable to capture the temporal relations of multiple images and fail to answer temporal reasoning questions in \ours{}. The existing LVLMs lack true temporal reasoning ability.
\end{enumerate}

We plan to expand this benchmark and figure out other ways to diagnose issues within LVLMs. 
We hope that \ours{} can be used to identify and provide insights on the weakness of different LVLMs, to facilitate finetuning and improvement of those models based on the diagnoses.


\section{Acknowledgements}
This research was supported by Army Cooperative Agreement W911NF2120076 and ARO  W911NF2310046 and W911NF2310352.
Our work is also supported in part by DARPA SemaFor Program under HR001120C0124. 
Zhou is supported in part by Adobe Research gift fund. 
Xiaoyu and Huang are supported by NSF-IIS-2147276 FAI, DOD N00014-22-1-2335 and FA9550-23-1-0048, DARPA GARD HR00112020007, Adobe, Capital One and JP Morgan.


\appendix

\section{More Case Analysis on \ours{} with GPT-4V and LLaVA-1.5}

In this section, we give a few samples in \ours{} and share our observations.
\textbf{Each figure is self-contained for readability}, where we highlight the control pairs, the responses of GPT-4V and LLaVA-1.5, the failures of those models, and the corresponding part of the answers. 

\subsection{Visual Dependent Examples}

From the famous illusions in Fig.\ref{visual_depend1}, Fig.\ref{visual_depend2}, and Fig.\ref{visual_depend3}, we found {GPT-4V is more knowledgeable than LLaVA-1.5} in recognizing all the illusion cases and knowing their names. However, GPT-4V fails to answer the question faithfully based on the edited images. The reason behind this might be that {GPT-4V tends to generate answers based on its parametric memory instead of analyzing the images}. Compared to GPT-4V, LLaVA-1.5 performs badly on both the original image and edited images, indicating that {the visual perception skill of LLaVA-1.5 is limited}. 

From the examples in Fig.\ref{visual_depend4} and Fig.\ref{visual_depend5}, we found both GPT-4V and LLaVA-1.5 are unable to correctly recognize parallel lines, regular triangles, polygons, and other math theorems, meaning that {geometry and math are still a challenging task for GPT-4V}. 

We further explore GPT-4V's and LLaVA-1.5's abilities in Optical Character Recognition in Fig.\ref{visual_depend6} and Figure Recognition in Fig.\ref{visual_depend7}. From our observations, we found that GPT-4V and LLaVA-1.5 are easily misled by editing the characters in the images, demonstrating that GPT-4V and LLaVA-1.5 generate answers based on their parametric memory instead of visual reasoning. This is because the difference between the original images and edited images is obvious. 

Inspired by~\cite{yang2023dawn}, which shows the promising video understanding of GPT-4V, we also investigate more examples in Fig.\ref{visual_depend_video1} and Fig.\ref{visual_depend_video2}, including several frame sequence examples. The positive sequence and reversed sequence have the opposite semantic meaning, such as \textit{"disappear or appear" and "park or leave"} in Fig.\ref{visual_depend_video1}. From the comparison, we found that {GPT-4V is unable to distinguish between the positive sequence and the reversed sequence of the images, indicating that there is still much room to improve the video reasoning ability}.

\subsection{Visual Supplement Examples}
In Fig.\ref{visual_supple1}, Fig.\ref{visual_supple2}, and Fig.\ref{visual_supple3}, GPT-4V does not have an affirmative answer if no images are given. Given the image context, GPT-4V and LLaVA-1.5 are unable to understand the chart correctly, indicating that {their chart reasoning ability is still limited}. In the second example (bottom) of Fig.\ref{visual_supple10}, the predictions of GPT-4V changed completely after we rotated the chart.

In Fig.\ref{visual_supple4}, 
Fig.\ref{visual_supple6}, Fig.\ref{visual_supple8}, Fig.\ref{visual_supple9}, and Fig.\ref{visual_supple10}, GPT-4V and LLaVA-1.5 have an affirmative answer if no images are given. After providing the image, including charts, tables, or maps, we found that they preferred to answer the questions with their knowledge instead of analyzing the image. This might be because {GPT-4V and LLaVA-1.5 demonstrate a marked dependence on textual reasoning capabilities, often prioritizing them over visual reasoning}. 

From Fig. \ref{visual_supple6} and Fig.\ref{visual_supple7}, we found the knowledge from LLaVA-1.5 is not accurate since it states "\textit{$\pi$ doesn't range from 3.1415926 and 3.1415927}" and "\textit{North Carolina is farther north than Delaware}." This observation also supports our claim that {GPT-4V is more knowledgeable than LLaVA-1.5}.

\section{Decision Tree Logic and Examples}

\definecolor{mymagenta}{RGB}{255,0,255}

\begin{figure*}[t]
    \centering
    \includegraphics[width=0.8\textwidth]{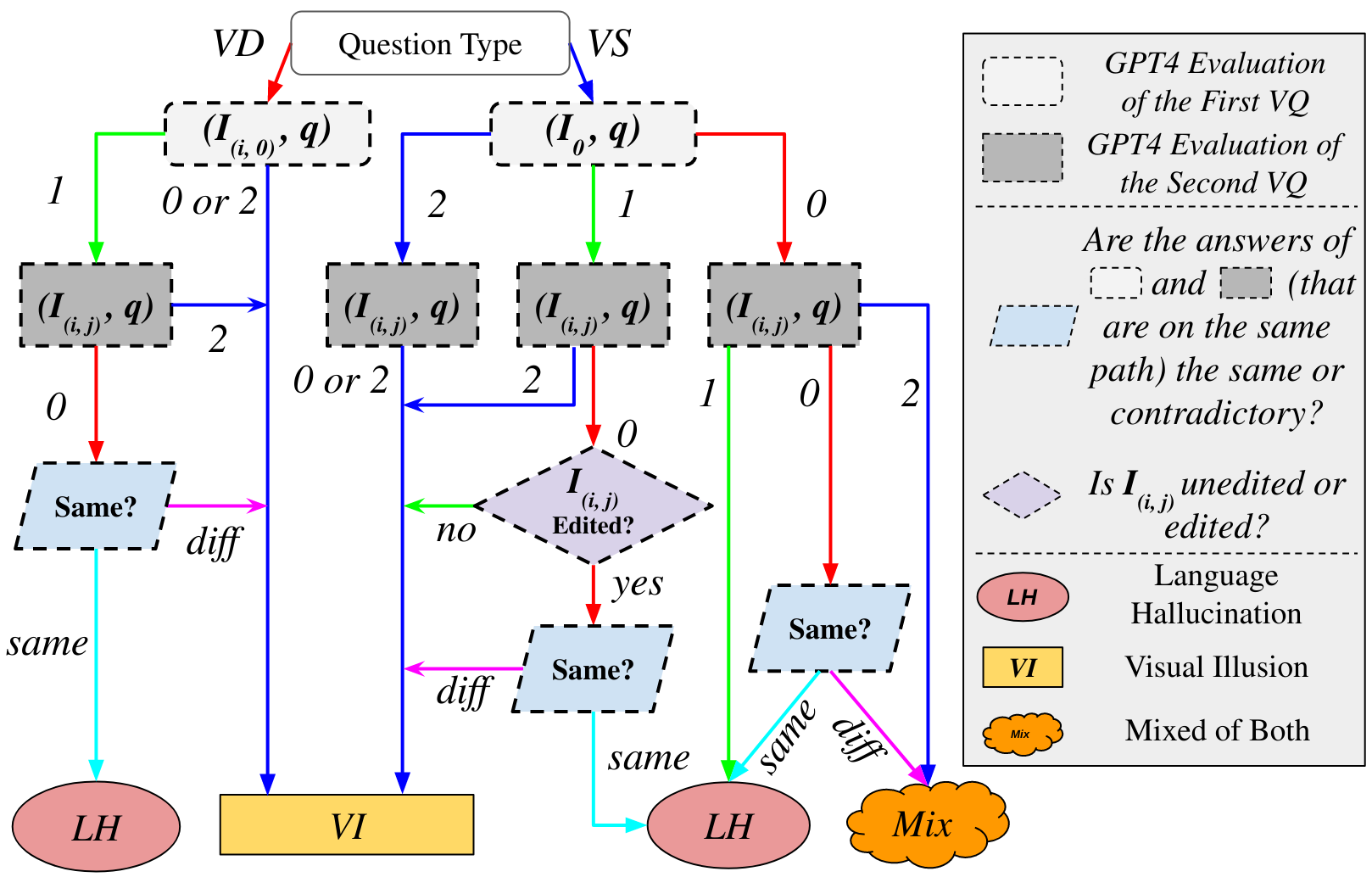}
    \caption{\textbf{Decision Tree to Diagnose Failure Types:} Based on the correctness of two questions in a control pair, and the difference in their responses, we use this decision tree to analyze the failure. We highlight different decision paths with \textcolor{red}{Red(R)}, \textcolor{blue}{Blue(B)}, \textcolor{green}{Green(G)}, \textcolor{cyan}{Cyan(C)} and \textcolor{mymagenta}{Magenta(M)}. So a path on the decision tree can be represented as a sequence of colors, e.g., \textcolor{red}{R}-\textcolor{green}{G}-\textcolor{red}{R}-\textcolor{cyan}{C}. The output of \textit{GPT4 Evalution} could be \textit{Incorrect (0)}, \textit{Correct (1)}, or \textit{Uncertain (2)} if the predicted response is ambiguous.}
    \label{fig:decisiontree_colored}
    \vspace{2mm}
     
\end{figure*}

In Fig.~\ref{fig:decisiontree_colored}, we utilize the decision tree to determine the failure types. In the rest of the section, specifically Fig.~\ref{a_tree_id_1}-\ref{a_tree_id_6}, we will provide a few examples and explain the logic that leads to different types of errors. \textbf{Each figure with its caption is self-contained for readability.}

In Fig.~\ref{a_tree_id_1} (bottom), it is a visual-dependent sample (VD). The answer regarding the original image is correct (1), but the answer to the edited image is incorrect (0), and the two answers are the same (\textit{same}). This shows that GPT-4V knows the \textit{"Chubb illusion"} in its parametric knowledge but can not answer according to the image. In Fig.~\ref{fig:decisiontree_colored}, these correspond to the (VD) \textcolor{red}{R}-\textcolor{green}{G}-\textcolor{red}{R}-\textcolor{cyan}{C} route in the decision tree, leading to the diagnostic result of \textit{Language Hallucination}. 

\begin{figure*}
\thisfloatpagestyle{plain}
\begin{center}
\begin{tikzpicture}
\node [mybox,title=Visual Dependent -- Decision Tree Logic] (box){%
    \begin{minipage}{2\columnwidth}

       \raggedright
        \bluebold{Decision Tree: }The \textcolor{red}{RED} path means the decision path.\\[2pt]
         \centering
        \includegraphics[width=0.6\columnwidth]{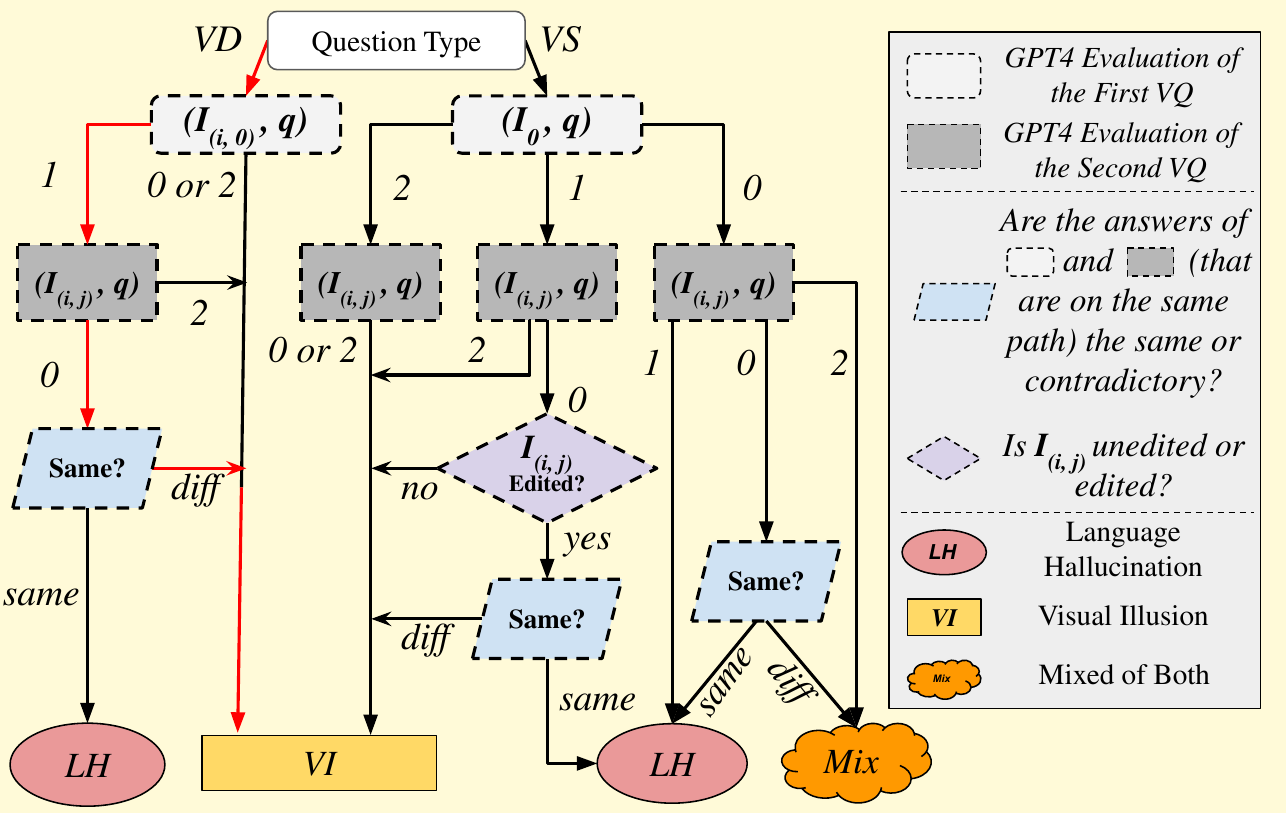}

        \raggedright
        \bluebold{Prompt: } Is the vertical line the same length as the horizontal line?
 \\[2pt]

    \begin{multicols}{2}

        \centering
        \includegraphics[width=0.4\columnwidth]{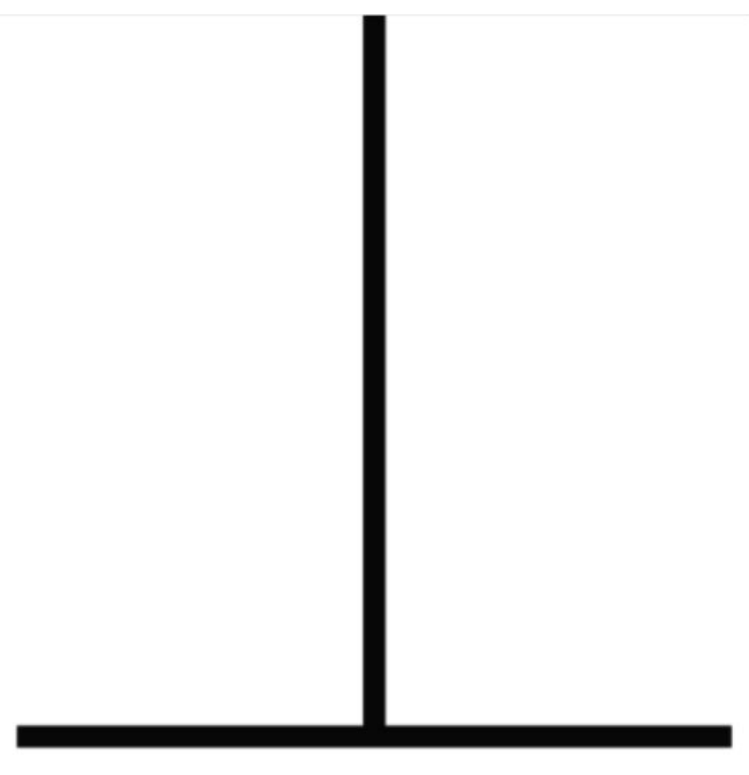}
        
        \textit{\textbf{Original:} The lines have the same length.} \\ [2pt]
        \raggedright
        \bluebold{Answer:} Yes, the lines have the same length.\\[10pt]

        \bluebold{GPT-4V: } Yes, the vertical line is the same length as the horizontal line.\\[2pt]

        \columnbreak
    
        \centering
        \includegraphics[width=0.4\columnwidth]{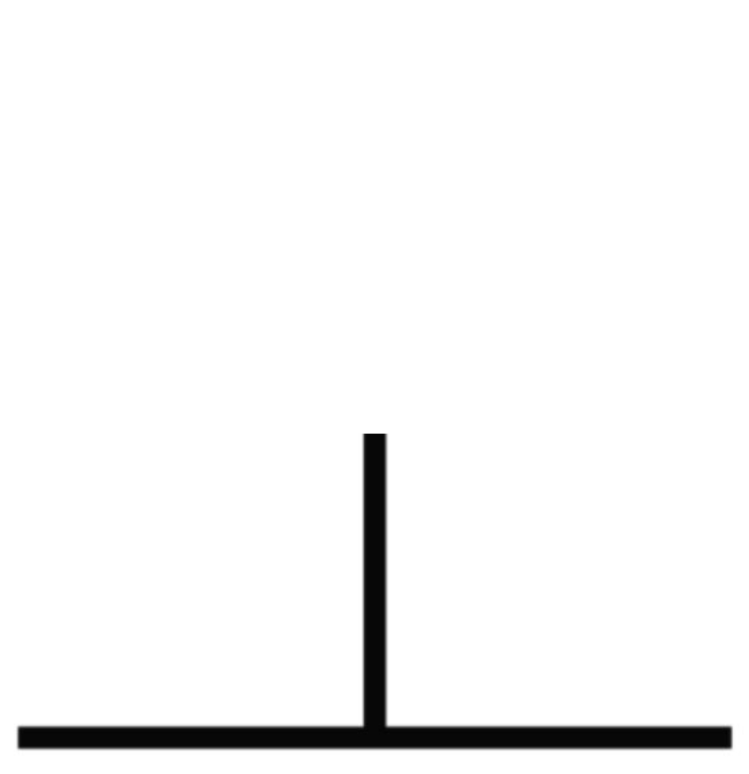}
        
        \textit{\textbf{Edited:} The vertical line is shortened.} \\ [2pt]
        
        \raggedright
        \bluebold{Answer:} No, the vertical line is shorter.\\[10pt]

        \bluebold{GPT-4V: } \illusion{No, the vertical line appears to be longer than the horizontal line in the image provided.} \\[2pt]
 
    \end{multicols}

    \end{minipage}
};
\end{tikzpicture}%
\caption{
It is a visual-dependent sample (VD). The answer regarding the original image is correct (1), but the answer to the edited image is incorrect (0), and the two answers are not the same. This shows that GPT-4V can not compare the length of the two lines correctly. This corresponds to the \textcolor{red}{RED} route in the decision tree, leading to the diagnostic result of \textit{Visual Illusion}. We highlight the \textcolor{red}{incorrect answer} according to \illusion{visual illusion}, \hallu{language hallucination}, or \mixed{potentially mixed}.
}
\label{a_tree_id_2}
\end{center}
\end{figure*}

In Fig.~\ref{a_tree_id_2} (bottom), it is a visual-dependent sample (VD). The answer regarding the original image is correct (1), but the answer to the edited image is incorrect (0), and the two answers are not the same (\textit{same}). This shows that GPT-4V can not compare the length of the two lines correctly. In Fig.~\ref{fig:decisiontree_colored}, it corresponds to the (VD) \textcolor{red}{R}-\textcolor{green}{G}-\textcolor{red}{R}-\textcolor{mymagenta}{M}-\textcolor{blue}{B} route in the decision tree, leading to the diagnostic result of \textit{Visual Illusion}. 

\begin{figure*}
\thisfloatpagestyle{plain}
\begin{center}
\begin{tikzpicture}
\node [mybox,title=Visual Dependent -- Decision Tree Logic] (box){%
    \begin{minipage}{2\columnwidth}

       \raggedright
        \bluebold{Decision Tree: }The \textcolor{red}{RED} path means the decision path.\\[2pt]
         \centering
        \includegraphics[width=0.6\columnwidth]{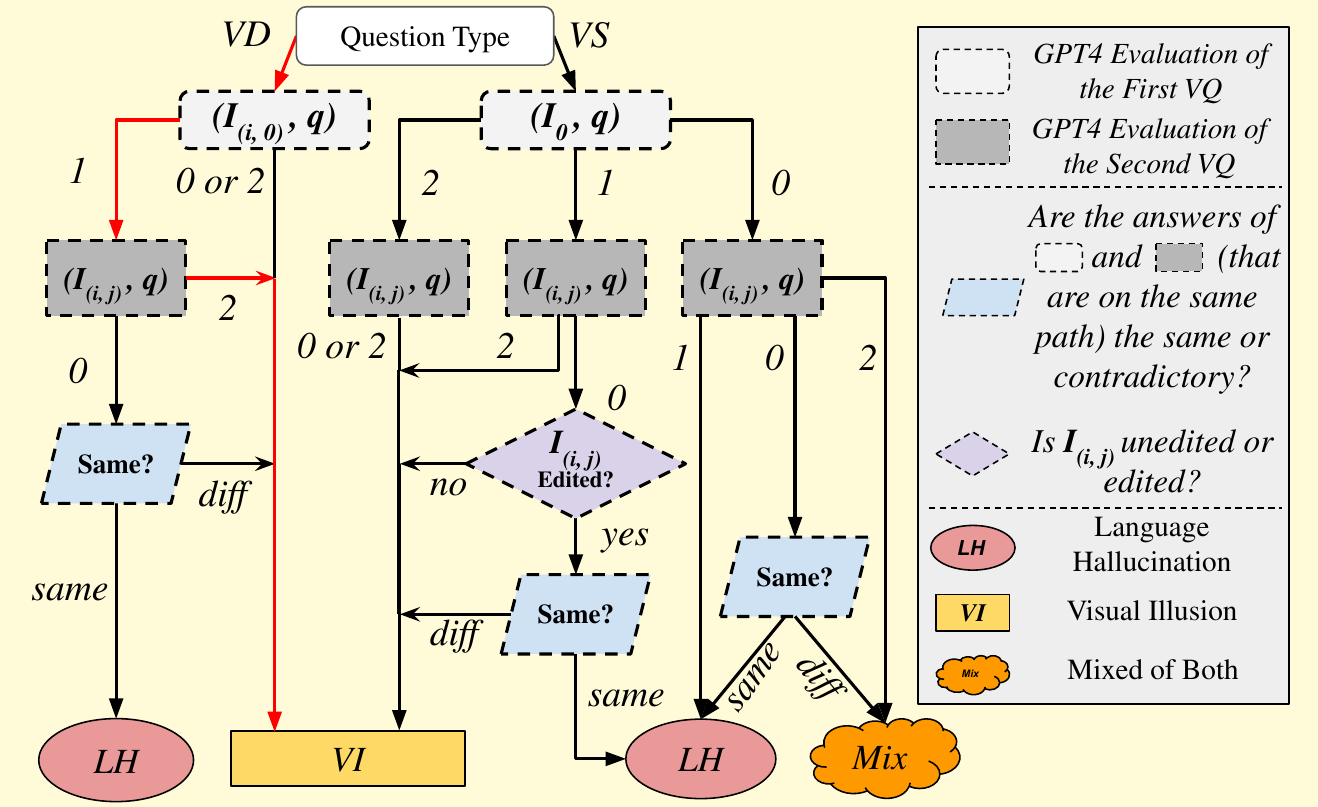}

        \raggedright
        \bluebold{Prompt: } Is the vertical line longer than the horizontal line?\\[2pt]

    \begin{multicols}{2}

        \centering
        \includegraphics[width=0.6\columnwidth]{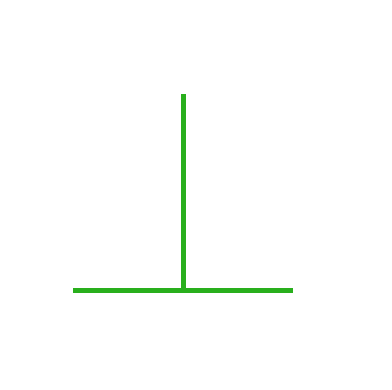}
        
        \textit{\textbf{Original image}} \\ [2 pt]
        \raggedright
        \bluebold{Answer:} No, the lines have the same length.\\[10pt]

        \bluebold{GPT-4V: } The vertical line is not longer than the horizontal line. \\[2pt]

        \columnbreak
    
        \centering
        \includegraphics[width=0.6\columnwidth]{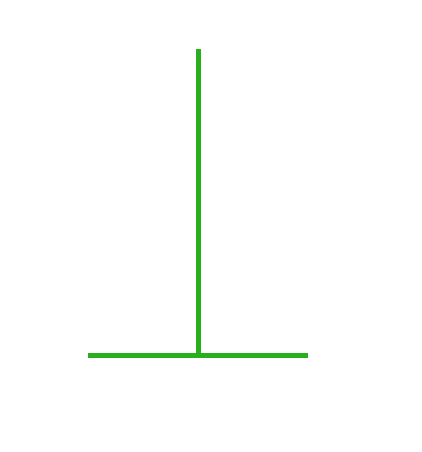}
        
        \textit{\textbf{Edited:} The vertical line is elongated.} \\ [2 pt]
        \raggedright
        \bluebold{Answer:} Yes, the vertical line is longer.\\[10pt]

        \bluebold{GPT-4V: } Without the ability to measure the lines, and based on common visual illusions, the lines may be of the same length. \hallu{However, this type of image is often used to demonstrate the vertical-horizontal illusion, where the vertical line appears longer than the horizontal line due to the way our perception works, even though they are the same length.} \illusionunsure{If this image is not intended to be an illusion and without exact measurement, I cannot definitively say whether one line is longer than the other.} \\[2pt]

    \end{multicols}

    \end{minipage}
};
\end{tikzpicture}%
\caption{
It is a visual-dependent sample (VD). The answer regarding the original image is correct (1), but the answer to the edited image is uncertain (2). This corresponds to the \textcolor{red}{RED} route in the decision tree, leading to the diagnostic result of \textit{Visual Illusion}.  We highlight the \textcolor{blue}{uncertain answer}, and the \textcolor{red}{incorrect answer} according to \illusion{visual illusion}, \hallu{language hallucination}, or \mixed{potentially mixed}. In this case, our decision tree can only detect \textit{visual illusion}; we can get additional information about the failure by examining the responses.
}
\label{a_tree_id_7}
\end{center}
\end{figure*}

In Fig.~\ref{a_tree_id_7} (bottom), it is a visual-dependent sample (VD). The answer regarding the original image is correct (1), but the answer to the edited image is uncertain (2). This shows that GPT-4V is uncertain about the length of the vertical line compared with the horizontal line. In Fig.~\ref{fig:decisiontree_colored}, it corresponds to the (VD) \textcolor{red}{R}-\textcolor{Green}{G}-\textcolor{blue}{B}-\textcolor{blue}{B} route in the decision tree, leading to the diagnostic result of \textit{Visual Illusion}.

In Fig.~\ref{a_tree_id_11} (bottom), It is a visual-dependent sample (VD). The answer regarding the original image is incorrect (0) or uncertain (2). This shows that LLaVA-1.5 fails to determine the diameters of the three circles in the original image, but succeeds in the edited image. In Fig.~\ref{fig:decisiontree_colored}, it corresponds to the (VS) \textcolor{red}{R}-\textcolor{blue}{B} route in the decision tree, leading to the diagnostic result of \textit{Visual Illusion}.

In Fig.~\ref{a_tree_id_3} (bottom), it is a visual-supplement sample (VS). The answer regarding the original image is uncertain (2), but the answer is incorrect (0) or uncertain (2) when the supplementary image is given. This shows that GPT-4V is uncertain about the answer without the visual input, and fails to answer the question with the supplementary image as well. In Fig.~\ref{fig:decisiontree_colored}, it corresponds to the (VS) \textcolor{blue}{B}-\textcolor{blue}{B}-\textcolor{blue}{B} route in the decision tree, leading to the diagnostic result of \textit{Visual Illusion}.

In Fig.~\ref{a_tree_id_12} (bottom), It is a visual-supplement sample (VS). The answer is correct (1) without being given any image. However, the answer is uncertain (2) when the supplementary image is given. This shows that GPT-4V is uncertain about the answer given the supplementary image though it could make the correct answer without the image. In Fig.~\ref{fig:decisiontree_colored}, it corresponds to the (VS) \textcolor{blue}{B}-\textcolor{Green}{G}-\textcolor{blue}{B}-\textcolor{blue}{B} route in the decision tree, leading to the diagnostic result of \textit{Visual Illusion}.

In Fig.~\ref{a_tree_id_9} (bottom), it is a visual-supplement sample (VS). The answer is already correct (1) without being given any image. However, the answer is incorrect (0) given the original supplementary image. The supplementary image is not edited. This shows that GPT-4V produces the wrong answer given the supplementary image, though it could produce the correct answer without the image. In Fig.~\ref{fig:decisiontree_colored}, it corresponds to the (VS) \textcolor{blue}{B}-\textcolor{Green}{G}-\textcolor{red}{R}-\textcolor{Green}{G}-\textcolor{blue}{B} route in the decision tree, leading to the diagnostic result of \textit{Visual Illusion}.

In Fig.~\ref{a_tree_id_10} (bottom), it is a visual-supplement sample (VS). The answer is correct (1) without being given any image. However, the answer is incorrect (0) when a edited image is given. The supplementary image is edited and the two answers are not the same. This shows that GPT-4V produces the wrong answer based on reasons inconsistent with the edited supplementary image, though it could produce a correct answer without the image. In Fig.~\ref{fig:decisiontree_colored}, it corresponds to the (VS) \textcolor{blue}{B}-\textcolor{Green}{G}-\textcolor{red}{R}-\textcolor{red}{R}-\textcolor{mymagenta}{M}-\textcolor{blue}{B} route in the decision tree, leading to the diagnostic result of \textit{Visual Illusion}.

In Fig.~\ref{a_tree_id_4} (bottom), it is a visual-supplement sample (VS). The answer is correct (1) without being given any image but the answer is incorrect (0) when an edited supplementary image is given. The supplementary image is edited by swapping Delaware and Arizona on the map. The two answers are the same. This indicates that GPT-4V has the prior knowledge of ``Delaware is the farthest north'' in its parametric knowledge but can not provide a correct answer according to the edited map. In Fig.~\ref{fig:decisiontree_colored}, it corresponds to the (VS) \textcolor{blue}{B}-\textcolor{Green}{G}-\textcolor{red}{R}-\textcolor{red}{R}-\textcolor{cyan}{C} route in the decision tree, leading to the diagnostic result of \textit{Language Hallucination}.

In Fig.~\ref{a_tree_id_8} (bottom), it is a visual-supplement sample (VS). The answer is incorrect (0) without being given any image. But the answer becomes correct given the original image. This indicates that LLaVA-1.5's answer is affected by hallucinations without given image information. In Fig.~\ref{fig:decisiontree_colored}, it corresponds to the (VS) \textcolor{blue}{B}-\textcolor{red}{R}-\textcolor{Green}{G} route in the decision tree, leading to the diagnostic result of \textit{Language Hallucination}.

In Fig.~\ref{a_tree_id_5} (bottom), it is a visual-supplement sample (VS). The answer is incorrect (0) without being given any image. The answer is still incorrect (0) when the original supplementary image is given. And the two answers are the same. This shows that LLaVA-1.5 has the issue of hallucinations with and without the image information. In Fig.~\ref{fig:decisiontree_colored}, it corresponds to the (VS) \textcolor{blue}{B}-\textcolor{red}{R}-\textcolor{red}{R}-\textcolor{cyan}{C} route in the decision tree, leading to the diagnostic result of \textit{Language Hallucination}.

In Fig.~\ref{a_tree_id_6} (bottom), it is a visual-supplement sample (VS). The answer is incorrect (0) without being given any image. The answer is still incorrect (0) when an edited supplementary image is given. However, the two answers are not the same. This indicates that the commonsense knowledge about the location of US states in LLaVA-1.5 is weak and wrong without the input image of the US map. Additionally, the visual interpretation of the map by LLaVA-1.5 is incorrect. In Fig.~\ref{fig:decisiontree_colored}, it corresponds to the (VS) \textcolor{blue}{B}-\textcolor{red}{R}-\textcolor{red}{R}-\textcolor{mymagenta}{M} route in the decision tree, leading to the diagnostic result of \textit{Potentially Mixed}.


\input{sup_figures/circle-id-1}

\input{sup_figures/arrow-id-3}
\input{sup_figures/illusion-id-5}
\input{sup_figures/parallel-id-7}

\input{sup_figures/angle-id-12}
\input{sup_figures/ad-id-15}
\input{sup_figures/cartoon-id-17}

\input{sup_figures/homer-video-1}
\input{sup_figures/plug-video-3}

\input{sup_figures/gold-is-1}
\input{sup_figures/bread-is-2}
\input{sup_figures/population-is-3}


\input{sup_figures/map-is-4}
\input{sup_figures/daleware-is-6}
\input{sup_figures/pai-is-7}
\input{sup_figures/population-is-8}
\input{sup_figures/olympic-is-9}
\input{sup_figures/chinaarea-is-10}

\input{sup_figures/a_tree_id_1}
\input{sup_figures/a_tree_id_11}
\input{sup_figures/a_tree_id_3}
\input{sup_figures/a_tree_id_12}
\input{sup_figures/a_tree_id_9}
\input{sup_figures/a_tree_id_10}
\input{sup_figures/a_tree_id_4}
\input{sup_figures/a_tree_id_8}
\input{sup_figures/a_tree_id_5}
\input{sup_figures/a_tree_id_6}

\begin{figure*}
\thisfloatpagestyle{plain}
\begin{center}
\begin{tikzpicture}
\node [mybox, title=Gemini Pro Vision Responses, inner ysep=10pt] (box){%
    \begin{minipage}{2\columnwidth}

    \begin{multicols}{2}

        \centering
        \includegraphics[width=0.6\columnwidth]{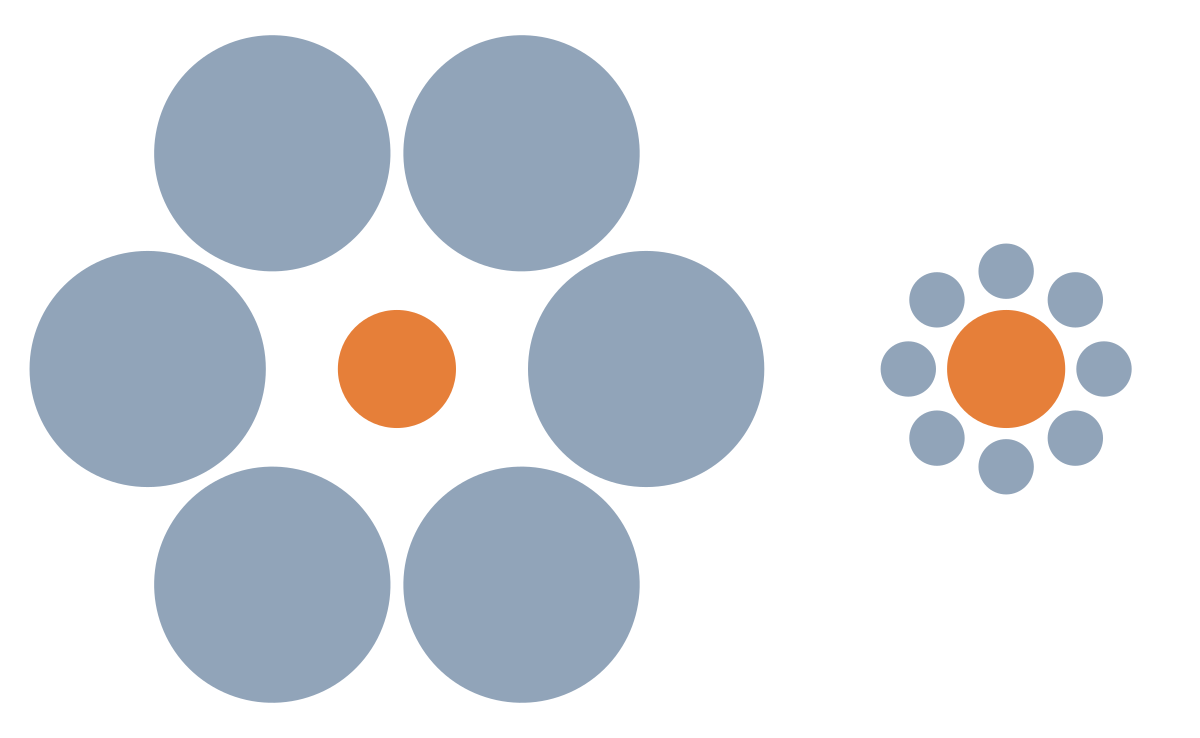}
        
        \textit{\textbf{Original:} Orange balls have the same size.} \\
        
        \columnbreak
    
        \centering
        \includegraphics[width=0.6\columnwidth]{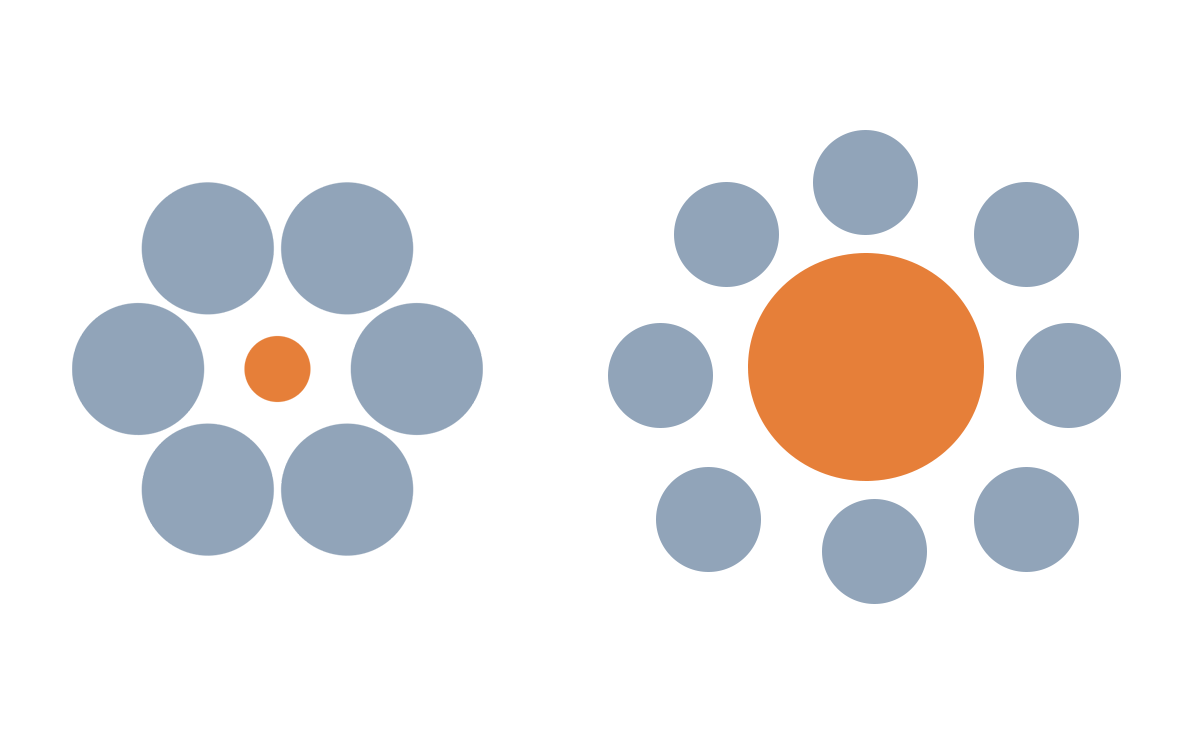}
        
        \textit{\textbf{Edited:} Orange balls have different size.} \\ 

    \end{multicols}

       \raggedright
        \bluebold{Prompt 1: } Is the right orange circle \ul{the same size} as the left orange circle? \\ \vspace{-10pt}

    \begin{multicols}{2}

        \centering

        \raggedright
        \bluebold{Answer:} Yes.\\[2pt]
        \bluebold{Gemini Pro: } Yes.

        \columnbreak
    
        \centering
        
        \raggedright
        \bluebold{Answer:} No, right orange circle is bigger.\\[2pt]
        \bluebold{Gemini Pro: } \textcolor{red}{Yes.}
    \end{multicols}

       \raggedright
        \bluebold{Prompt 2: } Is the right orange circle \ul{smaller than} the left orange circle? \\ \vspace{-10pt}
    \begin{multicols}{2}

        \centering

        \raggedright
        \bluebold{Answer:} No, they are of the same size.\\[2pt]
        \bluebold{Gemini Pro: } \textcolor{red}{Yes. \ul{The right orange circle is the sun. The left orange circle is closer to the camera.}}

        \columnbreak
    
        \centering
        
        \raggedright
        \bluebold{Answer:} No, right orange circle is bigger.\\[2pt]
        \bluebold{Gemini Pro: } \textcolor{red}{Yes.}
    \end{multicols}\vspace{-10pt}
            



    
        

        \par\noindent\rule[5pt]{\textwidth}{1pt}\\ \vspace{-15pt}

    \begin{multicols}{2}

        \centering
        \includegraphics[width=0.6\columnwidth]{raw_images/id_004_o.pdf}
        
        \textit{\textbf{Original:} The yellow lines have the same length.} \\ 
        
        \columnbreak
    
        \centering
        \includegraphics[width=0.6\columnwidth]{raw_images/id_004_e.pdf}
        
        \textit{\textbf{Edited:} The bottom yellow line is elongated.} \\ 

    \end{multicols}

        \raggedright
        \bluebold{Prompt: } Are the two yellow lines the same length? \\ \vspace{-10pt}

    \begin{multicols}{2}

        \centering
        \raggedright
        \bluebold{Answer:} Yes, the yellow lines have the same length.\\[2pt]
        \bluebold{Gemini Pro: } Yes. The two yellow lines are the same length.

        \columnbreak
    
        \centering
        
        \raggedright
        \bluebold{Answer:} No, the bottom yellow line is longer.\\[2pt]
        \bluebold{Gemini Pro: } \textcolor{red}{Yes. \ul{The two yellow lines are the same length. They are both the same distance from the woman's hand.}}

    \end{multicols}\vspace{-10pt}

        \par\noindent\rule[5pt]{\textwidth}{1pt}\\ \vspace{-15pt}

    \begin{multicols}{2}

        \centering
        \includegraphics[width=0.35\columnwidth]{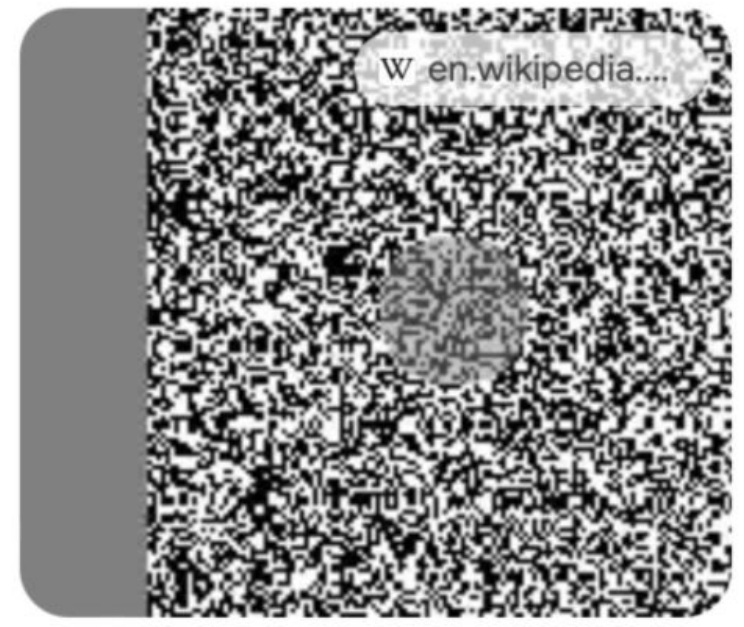}
        
        \textit{\textbf{Original:} The line in the middle is straight.} \\ 
        
        \columnbreak
    
        \centering
        \includegraphics[width=0.35\columnwidth]{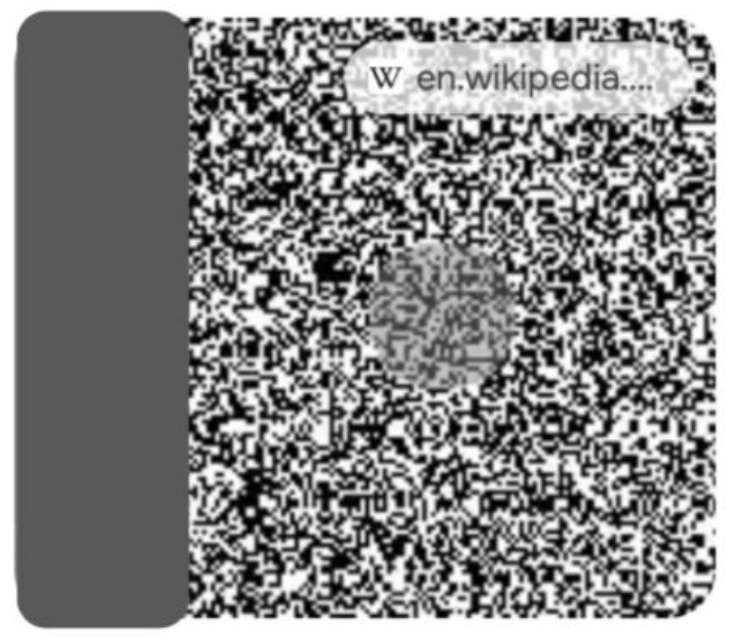}
        
        \textit{\textbf{Edited:} The line in the middle is curved.} \\ 

    \end{multicols}

        \raggedright
        \bluebold{Prompt: } Is the vertical line in the middle actually curved? \\ \vspace{-10pt}

    \begin{multicols}{2}

        \centering
        \raggedright
        \bluebold{Answer:} No, the line in the middle is straight.\\[2pt]
        \bluebold{Gemini Pro: } No. The vertical line in the middle is not curved. It is straight.

        \columnbreak
    
        \centering
        
        \raggedright
        \bluebold{Answer:} Yes, the line in the middle is curved.\\[2pt]
        \bluebold{Gemini Pro: } Yes, the line is curved. \textcolor{red}{\ul{The line is the edge of a wave. The waves in the ocean are curved.}}

    \end{multicols}

    \end{minipage}
};
\end{tikzpicture}%
\caption{
We highlight the \textcolor{red}{incorrect answer}.
}
\label{visual_depend1_gemini}
\end{center}
\end{figure*}

\clearpage

{
    \small
    \bibliographystyle{ieeenat_fullname}
    \bibliography{citation}
}


\end{document}